\documentclass{article}

\usepackage{microtype}
\usepackage{graphicx}
\usepackage{subfigure}
\usepackage{booktabs} 

\usepackage{hyperref}

\usepackage[accepted]{icml2024}

\usepackage{amsmath}
\usepackage{amssymb}
\usepackage{mathtools}
\usepackage{amsthm}

\usepackage{xcolor,colortbl}
\usepackage{marvosym}
\usepackage{wrapfig}
\usepackage{caption}
\usepackage{subfigure}
\usepackage{multirow}
\usepackage{url}

\usepackage{array}
\usepackage{booktabs}
\usepackage[title]{appendix}
\usepackage{verbatim}
\usepackage{algorithm}
\usepackage{algorithmic}
\usepackage{setspace}
\renewcommand{\algorithmicrequire}{\textbf{Input:}} 
\renewcommand{\algorithmicensure}{\textbf{Output:}}
\usepackage{rotating}

\usepackage{bbding}
\usepackage{threeparttable}
\usepackage{pifont}
\usepackage{arydshln}

\usepackage{graphicx}
\usepackage{listings}
\usepackage{tcolorbox}
\usepackage{utfsym}
\tcbuselibrary{skins}
\usepackage{makecell}
\usepackage{changepage}

\usepackage[capitalize,noabbrev]{cleveref}

\theoremstyle{plain}

\theoremstyle{definition}

\theoremstyle{remark}

\usepackage[textsize=tiny]{todonotes}

\newcommand{\ie}{\textit{i}.\textit{e}.}
\newcommand{\eg}{\textit{e}.\textit{g}.}

\begin{document}

\twocolumn[
\icmltitle{Accurate LoRA-Finetuning Quantization of LLMs via Information Retention}

\icmlsetsymbol{equal}{*}

\begin{icmlauthorlist}
\icmlauthor{Haotong Qin}{equal,eth}
\icmlauthor{Xudong Ma}{equal,buaa}
\icmlauthor{Xingyu Zheng}{buaa}
\icmlauthor{Xiaoyang Li}{bytedance}
\icmlauthor{Yang Zhang}{bytedance}
\icmlauthor{Shouda Liu}{bytedance}
\icmlauthor{Jie Luo}{buaa}
\icmlauthor{Xianglong Liu\textsuperscript{\Letter}}{buaa}
\icmlauthor{Michele Magno}{eth}
\end{icmlauthorlist}

\icmlaffiliation{buaa}{Beihang University}
\icmlaffiliation{eth}{ETH Z\"{u}rich}
\icmlaffiliation{bytedance}{Bytedance AI Lab}

\icmlcorrespondingauthor{\textsuperscript{\Letter}Xianglong Liu}{xlliu@buaa.edu.cn}

\icmlkeywords{Machine Learning, icml}

\vskip 0.3in
]

\printAffiliationsAndNotice{\icmlEqualContribution}

\begin{abstract}
The LoRA-finetuning quantization of LLMs has been extensively studied to obtain accurate yet compact LLMs for deployment on resource-constrained hardware. However, existing methods cause the quantized LLM to severely degrade and even fail to benefit from the finetuning of LoRA. This paper proposes a novel \textbf{IR-QLoRA} for pushing quantized LLMs with LoRA to be highly accurate through information retention. The proposed IR-QLoRA mainly relies on two technologies derived from the perspective of unified information: (1) statistics-based \textit{Information Calibration Quantization} allows the quantized parameters of LLM to retain original information accurately; (2) finetuning-based \textit{Information Elastic Connection} makes LoRA utilizes elastic representation transformation with diverse information. Comprehensive experiments show that IR-QLoRA can significantly improve accuracy across LLaMA and LLaMA2 families under 2-4 bit-widths, \eg, 4-bit LLaMA-7B achieves 1.4\% improvement on MMLU compared with the state-of-the-art methods. The significant performance gain requires only a tiny 0.31\% additional time consumption, revealing the satisfactory efficiency of our IR-QLoRA. We highlight that IR-QLoRA enjoys excellent versatility, compatible with various frameworks (\eg, NormalFloat and Integer quantization) and brings general accuracy gains. The code is available at \href{https://github.com/htqin/ir-qlora}{https://github.com/htqin/ir-qlora}.
\end{abstract}

\section{Introduction}
Large language models (LLMs) have demonstrated strong performance in natural language understanding~\cite{touvron2023LLaMA,touvron2023LLaMA2}. LLMs can be adapted to various downstream real-world applications, paired with large-scale pre-training and finetuning for downstream tasks~\cite{chang2023survey,devlin2018bert,zhao2023survey,huang2022towards,brown2020language}. However, because of the massive parameters and computation, the LLM has high or even harsh resource requirements for deployment scenarios. The inference of LLMs is expensive and heavily relies on high-performance devices, such as graphics processing units (GPUs)~\cite{ganesh2021compressing,zhu2023survey,chitty2023survey}. Therefore, compression approaches of LLMs are widely studied to allow their deployment on edge devices. 
Quantization emerges as a promising approach to compress LLMs by reducing bit-width but usually results in significant degeneration in accuracy~\cite{xiao2023smoothquant,lin2023awq}. 
For example, the 4-bit LLaMA-7B quantized by GPTQ~\cite{frantar2022gptq} suffers a 1.5\% drop of 5-shot accuracy on MMLU benchmark~\cite{hendrycks2020measuring} compared to its original counterpart~\cite{liu2023emergent}.

\begin{figure*}[t]
\centering
\includegraphics[width=1\textwidth]{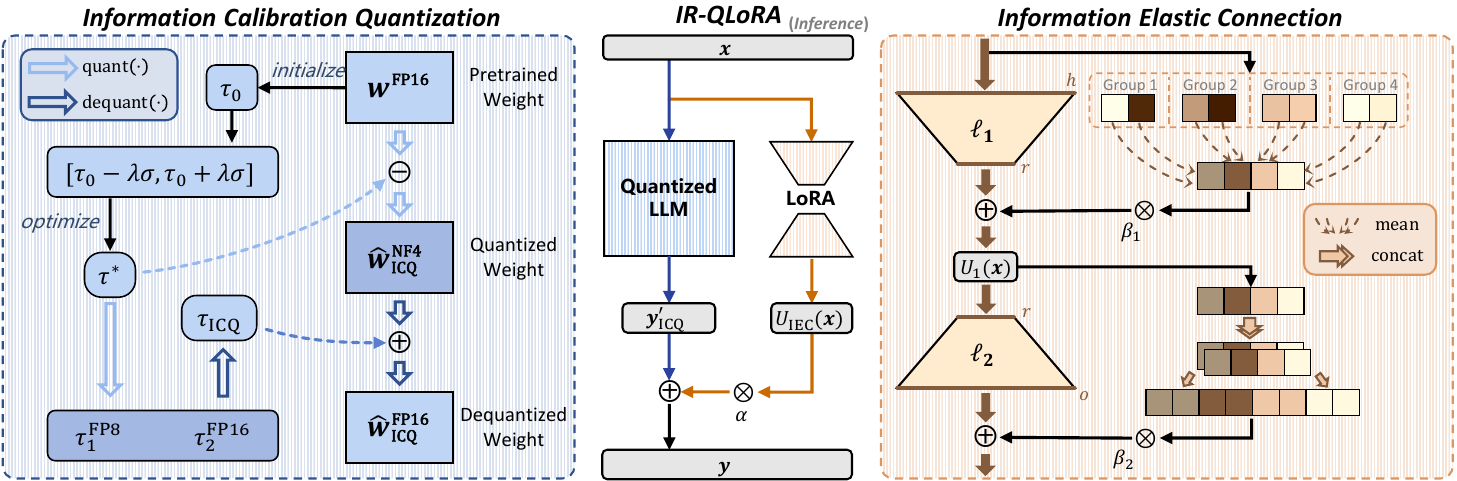}
\caption{Overview of IR-QLoRA. The framework includes Information Calibration Quantization
(ICQ) for quantizing LLMs and Information Elastic Connection (IEC) for enhancing LoRA} 
\label{fig:overview}
\end{figure*}

LoRA-finetuning quantization has become a popular paradigm that combines the LLM quantization with parameter-efficient finetuning of low-rank adaption (LoRA)~\cite{dettmers2023qlora,xu2023qalora}. Methods under this paradigm mainly consist of the following two phases. The first one is the post-training quantization (PTQ) of the LLM~\cite{dettmers20218}, obtaining quantizers by resource-saving calibration. The latter one is finetuning the LoRA~\cite{hu2021lora}, where the quantized LLM remains fixed and LoRA is finetuned. LoRA-finetuning quantization of LLMs is resource and time-saving compared to finetuning of the whole LLM while pushing the quantized LLM to high accuracy compared to performing PTQ solely~\cite{dettmers2023qlora,xu2023qalora,liu2023llm}.

However, despite several efforts made, existing LoRA fine-tuning quantization of LLMs is still far from the limits regarding accuracy. We empirically observe that the prevention of further accurate quantization is mainly because the information loss caused by LLM quantization is significant and cannot be recovered effectively by LoRA. Especially with ultra-low bit-widths ($\leqslant$ 3-bit) and large model scales ($\geqslant$ 30B), the former results in the nonlinearly increased level of information loss for each element, and the latter leads to a significant increase of the total amount of information loss for the whole model.
In these cases, the finetuned LoRA is hard to assist LLMs to achieve high accuracy on downstream tasks, \eg, 4-bit LLaMA-30B with finetuned LoRA even fails to achieve the accuracy of the original counterpart without finetuning (57.7\% \textit{vs.} 58.2\% on MMLU).

In this paper, we present \textbf{IR-QLoRA} to obtain accurate \textbf{Q}uantized LLMs with \textbf{LoRA} via \textbf{I}nformation \textbf{R}etention (see the overview in Figure~\ref{fig:overview}). To tackle the information loss of the quantization of the LLM, we propose a \textit{Information Calibration Quantization} (ICQ) technique. By calibration by entropy maximization, ICQ enables quantizers for the LLM to retain the original information from the original parameters to quantized ones. We also propose the \textit{Information Elastic Connection} (IEC) to enhance the information recovery capability of LoRA. IEC works together with LoRA, which performs parameter-free elastic transformations to utilize the information of original features and diversify the transformation form of LoRA.

Our IR-QLoRA provides strong and generic support to achieve accurate quantized LLMs with LoRA. Extensive experiments on the MMLU benchmark show that our IR-QLoRA outperforms existing methods with convincing margins on LLaMA and LLaMA2 series models under different bit-widths, especially at ultra-low bit-widths (2-3 bit). For example, the average accuracy of 2-bit IR-QLoRA in the LLaMA family is 0.5\% higher than SOTA LoRA-finetuning quantization methods. For efficiency, the significant performance growth brought by our IR-QLoRA requires only a tiny 0.31\% additional time consumption for LLaMA-13B. Moreover, IR-QLoRA is versatile and can boost existing LoRA-finetuning LLM quantization frameworks flexibly, \eg, the integration with QA-LoRA~\cite{xu2023qalora} brings a cost-free 0.5\% gain on MMLU to 4-bit LLaMA-7B.

\section{Related Work}

LLMs have demonstrated remarkable proficiency across diverse natural language understanding tasks and are established as a prominent paradigm in this field~\cite{chang2023survey,devlin2018bert,zhao2023survey,huang2022towards,brown2020language,touvron2023LLaMA,touvron2023LLaMA2}. This reality poses substantial challenges to deploying LLMs in settings with limited resources. Consequently, the research of the compression technologies for LLMs has gained prominence as a critical area of research.
Existing compression technologies of LLMs include pruning, distillation, low-rank decomposition, and low-bit quantization~\cite{ganesh2021compressing,zhu2023survey,chitty2023survey,xu2023tensorgpt}. Among these technologies, quantization aims to compress the LLMs from 16-bit floating-point to lower bit-widths to mitigate the storage and computation.

Since compression is from a generic bit-width perspective, quantization has become a popular method to obtain efficient LLMs~\cite{xiao2023smoothquant,lee2023owq,shao2023omniquant,dettmers2022llm,liu2023llm,kim2023memory}.
The LoRA-finetuning quantization of LLMs emerges to achieve a balanced trade-off between computational cost and accuracy~\cite{dettmers2023qlora,li2023loftq}, where quantized LLMs are finetuned with parameter-efficient LoRAs. However, existing quantized LLMs with LoRA are still far from ideal in accuracy.
More details about related works are presented in Appendix~\ref{sec:Details_about_Related_Works}.

\section{The Rise of IR-QLoRA}

\subsection{Preliminaries}
We first present a baseline for LoRA-finetuning quantization of LLMs following common practice~\cite{dettmers2023qlora}.

Before finetuning, the weights of LLMs are to be quantized. The quantization function for the weight $\boldsymbol{w}\in\mathbb{R}^{h \times o}$ is
\begin{equation}
\label{eq:generic_quant}
\hat{\boldsymbol{w}}^{\text{NF}k} ={\operatorname{NF}\!{k}}\left(\frac{\boldsymbol{w}}{s}\right) = {\operatorname{NF}\!{k}}\left(\frac{\boldsymbol{w}}{\operatorname{absmax}(\boldsymbol{w})}\right),
\end{equation}
where $\hat{\boldsymbol{w}}^{\text{NF}k}$ denotes quantized weight and the quantization block size is 64 as default, and $s$ is the scaling factor calculated by $\operatorname{absmax}(\boldsymbol{w})$. 
${\text{NF}{k}}(\cdot)$ denotes the $k$-bit NormalFloat quantization~\cite{dettmers2023qlora}, quantizing the weights of LLMs to $2^k$ values $q_i$ as follows:
\begin{equation}
q_i=\frac{1}{2}\left(Q\left(\frac{i}{2^k+1}\right)+Q\left(\frac{i+1}{2^k+1}\right)\right),
\end{equation} 
where $Q(\cdot)$ is the quantile function of $\mathcal{N}(0, 1)$ distribution.
Then, the computation process (\eg, linear projection) of the quantized unit of the LLM during inference is
\begin{equation}
\label{eq:projection}
\boldsymbol{y}'=\boldsymbol{x}\hat{\boldsymbol{w}}^\text{FP16}=\boldsymbol{x}\left(\hat{\boldsymbol{w}}^{\text{NF}k}\operatorname{dequant}(s_1^{\text{FP8}}, s_2^{\text{FP16}})\right),
\end{equation}
where $\boldsymbol{x}\in\mathbb{R}^{b \times h}$ and $\boldsymbol{y}'\in\mathbb{R}^{b \times o}$ denote the input and output of quantized linear projection in LLMs, respectively.
$\operatorname{dequant}(s_1^{\text{FP8}}, s_2^{\text{FP16}})$ is expected to approximate the original scaling factor $s$. After double-quantization of $s$, we can obtain the quantized values $s_1^{\text{FP8}}$ and scaling factors $s_2^{\text{FP16}}$ follow~\cite{dettmers2023qlora}.
$\hat{\boldsymbol{w}}^\text{FP16}$ denotes the FP16 weights dequantized from $\hat{\boldsymbol{w}}^{\text{NF}k}$.

\begin{figure}[t]
\centering
\includegraphics[width=1\linewidth]{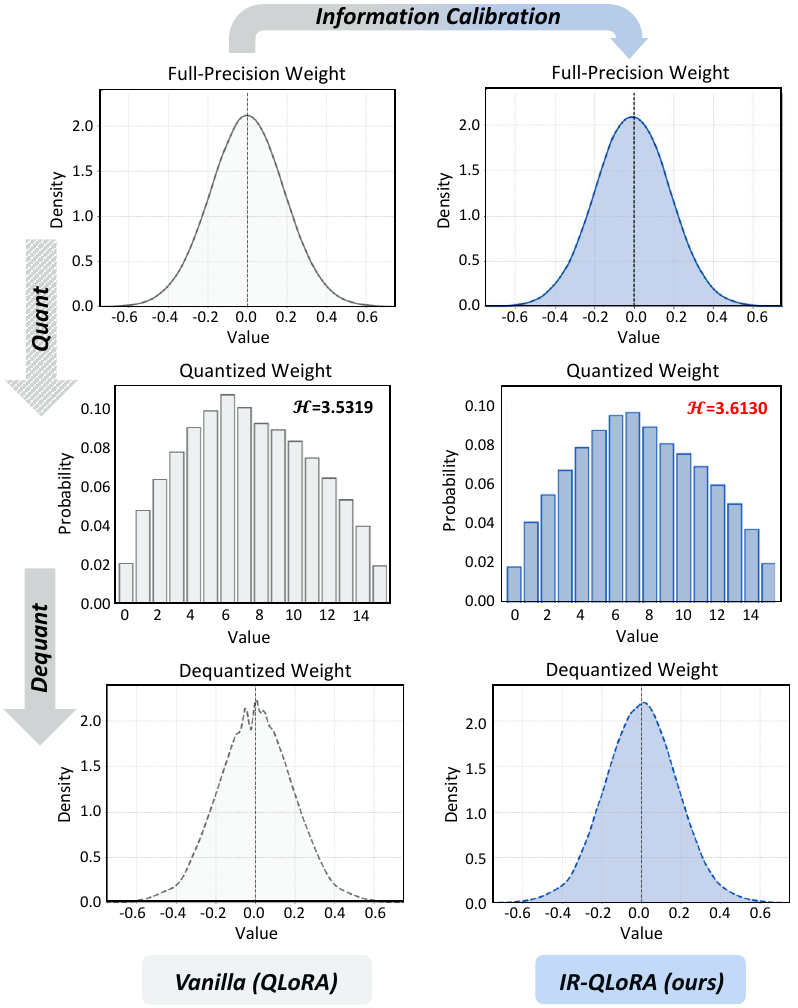}
\caption{An illustration of ICQ in IR-QLoRA} 
\label{fig:method-ICQ}
\end{figure}

The LoRA refers to a set of finetunable parameters designed to enhance the quantized linear projection in LLMs by introducing an extra factorized projection~\cite{hu2021lora,dettmers2023qlora}. For the quantized linear projection as Eq.~\eqref{eq:projection}, the computation with LoRA can be expressed as:
\begin{equation}
\label{eq:lora}
\boldsymbol{y} = \boldsymbol{y}' + \alpha \boldsymbol{x}\boldsymbol{\ell}_1\boldsymbol{\ell}_2,
\end{equation}
where $\boldsymbol{\ell}_1\in\mathbb{R}^{h \times r}$ and $\boldsymbol{\ell}_2\in\mathbb{R}^{r \times o}$ are the finetunable parameters, and $\alpha$ is a scalar.
Since the parameter efficiency of LoRA should be kept during inference, its rank $r$ is far smaller than the input and output dimensions ($h$ and $o$, respectively), which makes its memory and computational consumption far smaller than the corresponding linear projection in LLMs (\eg, $r=64$ \textit{vs.} $h=4096$ and $o=4096$). During the backward propagation of the finetuning process, the gradients are passed through the fixed quantized weights of LLMs to update the parameters in LoRA.

The quantization process of the LLM and the finetuning process of the LoRA are decoupled. The PTQ first processes the LLM to obtain low-bit quantized weights, and then the LoRA is finetuned for specific downstream tasks.

\subsection{Information Calibration Quantization}

\subsubsection{Degeneration of Quantized LLMs}
In the aforementioned baseline, the LLMs are quantized directly from pre-trained models, where the low-bit discretization of the parameters causes the accuracy degradation. Existing quantization methods attribute the degradation to the numerical quantization error. However, the information loss caused by quantization is always neglected. 

Specifically, the quantized weights of LLMs are expected to reflect the information carried by original counterparts, but reduced bit-width severely constrains the representation capabilities. 
From the information perspective, the dependence between the weights of quantized and original LLMs is expressed as the mutual information~\cite{qin2023distribution}:
\begin{equation}
\label{eq:im}
\mathcal{I}(\hat{\boldsymbol{w}}^{\text{FP16}}; \boldsymbol{w})
= \mathcal{H}(\hat{\boldsymbol{w}}^{\text{FP16}}) - \mathcal{H}(\hat{\boldsymbol{w}}^\text{FP16}\mid \boldsymbol{w}),
\end{equation}
where ${\mathcal{H} (\hat{\boldsymbol{w}}^{\text{FP16}})}$ denotes the entropy of $\hat{\boldsymbol{w}}^{\text{FP16}}$, and ${\mathcal {H} (\hat{\boldsymbol{w}}^\text{FP16}\mid \boldsymbol{w})}$ denotes the conditional entropy of $\hat{\boldsymbol{w}}^{\text{FP16}}$ given $\boldsymbol{w}$. As deterministic quantizers are used in the quantization of LLMs, $\mathcal{H}(\hat{\boldsymbol{w}}^{\text{FP16}}\mid \boldsymbol{w})=0$ and the $\mathcal{I}(\hat{\boldsymbol{w}}^{\text{FP16}}; \boldsymbol{w})$ depends on $\mathcal{H}(\hat{\boldsymbol{w}}^{\text{FP16}})$ directly.
In the PTQ, since the original weights $\boldsymbol{w}$ remain unchanged, maximizing the mutual information $\mathcal{I}(\hat{\boldsymbol{w}}^{\text{FP16}}; \boldsymbol{w})$ in Eq.~\eqref{eq:im} is equivalent to
\begin{equation}
\label{eq:ie}
\mathop{argmax}\limits_{s, s_1^{\text{FP8}}, s_2^{\text{FP16}}}\mathcal{H}(\hat{\boldsymbol{w}}^{\text{FP16}}; s, s_1^{\text{FP8}}, s_2^{\text{FP16}}).
\end{equation}
Since $\operatorname{dequant}(s_1^{\text{FP8}}, s_2^{\text{FP16}})$ is a scalar in dequantization and does not affect information entropy of $\hat{\boldsymbol{w}}^\text{FP16}$, the above objective function can be further simplified as follows:
\begin{equation}
\label{eq:ie_1}
\mathop{argmax}\limits_{s}\mathcal{H}(\hat{\boldsymbol{w}}^{\text{NF}k}; s)
= -\sum_{i=1}^{2^k-1} P(q_i) \log_{2} P(q_i),
\end{equation}
where $P(q_i)$ is the probability of $\hat{\boldsymbol{w}}^{\text{NF}k}$ taking the value $q_i$.

Since the significant reduction of bit-width leads to decreased representation capability, the entropy of the quantized weight is far less than that of the original counterpart. 
For example, the number of representation candidates for a 4-bit quantized weight reduces $4096\times$ compared to its original 16-bit (FP16) counterpart, and the upper bound of information entropy $\mathcal{H}(\hat{\boldsymbol{w}}^\text{FP16})$ in Eq.~\eqref{eq:ie} is correspondingly reduced $4\times$ (4 for 4-bit \textit{vs.} 16 for 16-bit), meaning a significant degradation of information in the quantity and quality.
Thus, prioritizing information recovery within low-bit weights is crucial for enhancing quantized LLMs.

\subsubsection{Information Calibration Quantization for Representation Recovery}
To mitigate the degeneration caused by information loss for the quantized LLMs, we introduce an Information Calibration Quantization (ICQ) technique for LLMs (as Figure~\ref{fig:method-ICQ}), which applies the fine-grained information maximization of quantized weights to improve the accuracy.

We first introduce a calibration constant $\tau$ to the quantizer, liberating its flexibility to retain information fully. The quantization process it engages in can be expressed as follows:
\begin{equation}
\label{eq:generic_quant_icq}
\hat{\boldsymbol{w}}^{\text{NF}k}={\text{NF}{k}}\left(\frac{\boldsymbol{w}-\tau}{s}\right).
\end{equation}
Since the original weight $\boldsymbol{w}$ is fixed, the optimization objective in Eq.~\eqref{eq:ie} can be expressed as
\begin{equation}
\label{eq:obj-1}
\mathop{argmax}\limits_{\tau, s}\mathcal{H}\left(\hat{\boldsymbol{w}}^{\text{NF}k}; \tau, s\right).
\end{equation}
Directly solving the objective in Eq.~\eqref{eq:obj-1} is significantly complex and time-consuming. Therefore, we then present a two-step strategy for calibrating the quantizers of LLMs blockwisely by the information entropy maximization.

\textbf{The first step} is to process the initialization for the calibration constant $\tau$. Based on the common assumption of a symmetrical normal distribution for weights of neural networks~\cite{dettmers2023qlora,baskin2021uniq}, we initialize the constant as the median $\tau_0=\operatorname{quantile}_{\frac{1}{2}}\left(\boldsymbol{w}\right)$ for each quantization block of weights. As the probability density is higher in regions closer to the symmetry axis in a normal distribution, this initialization enables the quantizer to utilize intervals to a greater extent. The position-dependent median also allows $\tau_0$ to alleviate the influence from outliers.

\textbf{The second step} is to optimize the calibration constant $\tau$ and the scaling factors $s_1$ and $s_2$ for quantization and double quantization of weights, respectively. We apply information entropy in Eq.~\eqref{eq:im} as the metric and perform the search-based optimization to obtain $\tau^*$. We create the search spaces for $\tau$ by linearly dividing [$\tau_0 - \lambda\sigma$, $\tau_0 + \lambda\sigma$] into $2n$ candidates, where $\sigma = 1$ is the standard deviation of $\mathcal{N}(0, 1)$ and $\lambda$ is a coefficient. We empirically set $\lambda$ and $n$ to 0.1 and 100, respectively, to achieve a trade-off between accurate and efficient search.
For each candidate $\tau$, we use it to calibrate weight and calculate the information entropy of weights quantized by Eq.~\eqref{eq:generic_quant_icq}, and then obtain the optimal calibration constant $\tau^*$ corresponding to the maximum entropy.
The scaling factor $s$ for $\boldsymbol{w}-\tau^*$ is double-quantized to $s_1^\text{FP8}$ and $s_2^\text{FP16}$.

For the optimized calibration constant $\tau^*$, we perform double quantization similar to the scale to save memory. 
The quantization and dequantization processes of our ICQ can be summarized as
\begin{equation}
\begin{aligned}
&\hat{\boldsymbol{w}}_{\text{ICQ}}^{\text{NF}k}={\text{NF}{k}}\left(\frac{\boldsymbol{w}-\tau^*}{\operatorname{absmax}(\boldsymbol{w}-\tau^*)}\right),\\
&\hat{\boldsymbol{w}}_{\text{ICQ}}^\text{FP16}=\hat{\boldsymbol{w}}_{\text{ICQ}}^{\text{NF}k}\operatorname{dequant}(s_1^{\text{FP8}}, s_2^{\text{FP16}})+\operatorname{dequant}(\tau_1^{\text{FP8}}, \tau_2^{\text{FP16}}).
\end{aligned}
\end{equation}
where $\tau_1^{\text{FP8}}$ and $\tau_2^{\text{FP16}}$ are the quantized calibration constant and its scaling factor for double quantization, respectively.
The inference of LLMs with our ICQ can be expressed as
\begin{equation}
\boldsymbol{y}'_\text{ICQ}=\boldsymbol{x}\hat{\boldsymbol{w}}_{\text{ICQ}}^{\text{FP16}}.
\end{equation}

Our ICQ technique maximizes the information entropy of quantized weight to alleviate its immense information degradation and revive the representation capability. As Figure~\ref{fig:method-ICQ} shows, the quantized weights calibrated by ICQ derive increased information and recover the original distribution more accurately following the dequantization.

\subsection{Information Elastic Connection}

\subsubsection{Limitation of Finetunable LoRA}
In addition to the quantized LLM in the baseline, the limited representation capability of the finetuneable LoRA also hinders information recovery. LoRA mitigates the performance degradation caused by weight quantization in LLM by finetuning additional adapters for downstream tasks, sometimes even yielding notable performance enhancements. The finetuning of LoRA can be roughly considered as finetuning a subset of weights in LLM, where low-rank parameters facilitate an efficient finetuning process, avoiding the expensive computation and storage of finetuning the LLM directly.

However, through LoRA processes parameter-efficient finetuning for quantized LLMs, its information representation still exhibits significant limitations, impeding the accurate quantized LLM with LoRA. Firstly, compared to the corresponding linear projection in LLM, the parameter transformation by LoRA can still be considered homogenized, albeit with a noticeably lower rank. On the other hand, the information utilization of LoRA remains limited, as the latter low-rank matrix $\boldsymbol{\ell}_2$ in Eq.~\eqref{eq:projection} can only exploit the transformed representation matrix $\boldsymbol{\ell}_1$ from the preceding transformation, thereby losing accessibility to the original representation information. Therefore, liberating LoRA from its constraints in representation capacity is expected to enhance the accuracy of quantized LLM further.

\begin{figure}[t]
\centering
\includegraphics[width=1\linewidth]{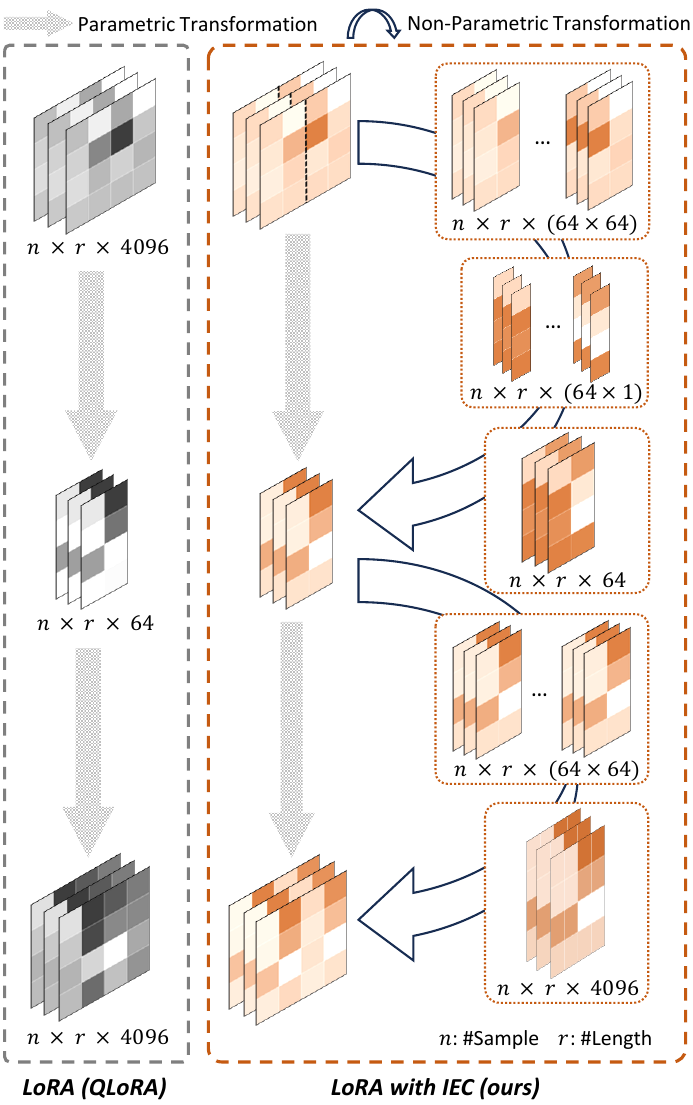}
\caption{An illustration of IEC in IR-QLoRA}
\label{fig:method-IEC}
\end{figure}

\subsubsection{Information Elastic Connection for Information Enhancement}

To bolster the representation capacity of LoRA, aiding in recovering information of quantized LLMs while maintaining its lightweight nature, we introduce an effective Information Elastic Connection (IEC). As Figure~\ref{fig:method-IEC} shows, IEC constructs a parameter-free connection for LoRA, facilitating information utilization derived from quantized LLM and diversifying the information transformation.

As shown in Eq.~\eqref{eq:lora}, the input of one LoRA unit is generated by the previous quantized LLM and LoRA units. The $h$-dimensional input $\boldsymbol{x}$ is transformed to low-rank $r$-dimensional intermediate features through the $\boldsymbol{\ell}_1$ matrix and then restored to the $o$-dimensional output through the $\boldsymbol{\ell}_2$ matrix.
Since usually $r \ll \min\{h, o\}$, we construct flexible parameter-free connections for LoRA matrix pairs so that each LoRA matrix can fully utilize the original input representation $\boldsymbol{x}$.
Specifically, we group and average the original feature according to the greatest common divisor of the input and intermediate dimensions and add it to the output computed by the $\boldsymbol{\ell}_1$ matrix. The first sub-unit $U_1$ of LoRA with our IEC can be expressed as
\begin{equation}
\label{eq:IEC_eq1}
\begin{aligned}
&U_1(\boldsymbol{x})=\boldsymbol{x}\boldsymbol{\ell}_1+\\
&\beta_1\prod\limits^{\frac{r}{\operatorname{gcd}{(h, r)}}}\left(\frac{\operatorname{gcd}{(h, r)}}{h}\sum\limits_{i=1}^{\operatorname{gcd}{(h, r)}}\boldsymbol{x}^{\left[(i-1)\frac{h}{\operatorname{gcd}{(h, r)}} : i\frac{h}{\operatorname{gcd}{(h, r)}}-1\right]}\right),
\end{aligned}
\end{equation}
where $\boldsymbol{x}^{\left[m: n\right]}$ denotes taking the $m$ to $n$ dimension of $\boldsymbol x$, $\operatorname{gcd}{(h, r)}$ denotes the greatest common divisor of $h$ and $r$, and $\beta$ denotes a layerwise learnable scalar; $\sum$ denotes the summation of all divided features and $\prod$ denotes the $\frac{r}{\operatorname{gcd}{(h, r)}}$-time repeated concatenation; The dimension of the latter term in Eq.~\eqref{eq:IEC_eq1} is $r$. Through the above operations, we transform the representation to low-rank through parameter-free operation to retain the original information.

The latter matrix of LoRA transforms the low-rank intermediate representation up to the higher dimension. It thus accompanies a parameter-free repeated concatenation for $\boldsymbol{x}'=U_1(\boldsymbol{x})$ and composes the second sub-unit $U_2$ in IEC, where the computation process can be expressed as
\begin{equation}
\label{eq:IEC_eq2}
\begin{aligned}
&U_2(\boldsymbol{x}')=\boldsymbol{x}'\boldsymbol{\ell}_2+\\
&\beta_2\prod\limits^{\frac{o}{\operatorname{gcd}(o, r)}}\left(\frac{\operatorname{gcd}(o, r)}{r}\sum\limits_{i=1}^{\operatorname{gcd}(o, r)}{\boldsymbol{x}'}^{\left[(i-1)\frac{r}{\operatorname{gcd}(o, r)}: i\frac{r}{\operatorname{gcd}(o, r)}-1\right]}\right),
\end{aligned}
\end{equation}
where $\boldsymbol{x}'$ is first aligned to the $\operatorname{gcd}(o, r)$-dimension and concatenated $\frac{o}{\operatorname{gcd}(o, r)}$ times repeatedly, and is then connected to the calculation result of $\boldsymbol{\ell}_2$ matrix.

Note that in the LoRA unit, the input dimension $h$ and output dimension $o$ are usually multiples of low-rank $r$, and $\operatorname{gcd}(h, r)$ and $\operatorname{gcd}(o, r)$ are thus equal to $r$. In these cases, Eq.~\eqref{eq:IEC_eq1} and Eq.~\eqref{eq:IEC_eq2} for our IEC can be simplified as follows:
\begin{equation}
\label{eq:IEC_eq}
\begin{aligned}
&U_1(\boldsymbol{x})=\boldsymbol{x}\boldsymbol{\ell}_1+
\beta_1\frac{r}{h}\sum\limits_{i=1}^{r}\boldsymbol{x}^{\left[(i-1)\frac{h}{r}: i\frac{h}{r}-1\right]},\\
&U_2(\boldsymbol{x}')=\boldsymbol{x}'\boldsymbol{\ell}_2+\beta_2\prod\limits^{\frac{o}{r}}{\boldsymbol{x}'}.
\end{aligned}
\end{equation}
With our IEC, the computation process of the quantized LLM projection and LoRA can be expressed as
\begin{equation}
\begin{aligned}
\boldsymbol{y} = \boldsymbol{y}'_\text{ICQ} + \alpha U_\text{IEC} (\boldsymbol{x}) =\boldsymbol{y}'_\text{ICQ} + \alpha U_2 \circ U_1 (\boldsymbol{x}).
\end{aligned}
\end{equation}
Appendix~\ref{Details about Pipeline} discusses the efficiency of IEC, which can be merged into LoRA to avoid additional inference costs.

The IEC propagates the input with elasticity dimension changing, thus allowing the matrix in LoRA to directly access and utilize the original information extracted by the quantized LLM projection. Moreover, the parameter-free IEC can seem diversified compared with the parametric matrix multiplication of LoRA, further enhancing the information representation of quantized LLMs.

\section{Experiment}
We extensively evaluate the accuracy and efficiency of our proposed IR-QLoRA.
Our IR-QLoRA is established upon the LLaMA~\cite{touvron2023LLaMA} and LLaMA2~\cite{touvron2023LLaMA2} families (7B, 13B, 30B, and 65B), and constructs parameter-efficient finetuning on Alpaca~\cite{taori2023stanford} and Flan v2~\cite{longpre2023flan} datasets. The Massively Multitask Language Understanding (MMLU)~\cite{hendrycks2020measuring} and CommonsenseQA benchmarks(\eg HellaSwag~\cite{zellers2019hellaswag}, PIQA~\cite{bisk2020piqa}) are applied for evaluation. For experiment settings~\cite{dettmers2023qlora,xu2023qalora,kim2023memory,frantar2022gptq}, we follow the settings of comparison methods reported in their publications or official code for fair comparison. All our experiments are conducted on Nvidia Tesla A100 GPUs.
Detailed experiment settings and results are presented in Appendix~\ref{sec:Detailed_Experiment_Settings} and Appendix~\ref{sec:Detailed_Experiment_Results}, respectively.

\begin{table}[t]
    \centering
    \caption{Accuracy (\%) comparison of LLaMA on the MMLU finetuned on the Alpaca dataset}
    \vspace{-0.1in}
    \label{tab:mmlu_alpaca}
    \setlength{\tabcolsep}{.7mm}
    {
    \begin{tabular}{lcccccc}
    \toprule
        \multirow{2}{*}{\textbf{Method}} & \multirow{2}{*}{\textbf{\#Bit}} & \multicolumn{5}{c}{\textbf{MMLU}} \\ \cmidrule(lr{.75em}){3-7} 
        ~ & ~ & \textbf{Hums.} & \textbf{STEM} & \textbf{Social} & \textbf{Other} & \textbf{Avg.}  \\ \midrule
        LLaMA-7B & 16 & 33.3 & 29.8 & 37.8 & 38.0 & 34.6  \\
        \hdashline[0.5pt/2pt]
        PEQA &  4 & 34.9 & 28.9 & 37.5 & 40.1 & 34.8  \\ 
        NormalFloat & 4 & 33.1 & 30.6 & 38.8 & 38.8 & 35.1 \\
        QLoRA {\scriptsize w/ GPTQ} & 4 & 33.8 & 31.3 & 37.4 & 42.2 & 36.0  \\
        QLoRA &  4 & 36.1 & 31.9 & 42.0 & 44.5 & 38.4  \\ 
        QA-LoRA &  4 & 36.6 & 32.4 & 44.8 & 44.9 & 39.4  \\ 
        \textbf{IR-QLoRA} & 4 & 38.6 & 34.6 & 45.2 & 45.5 & \textbf{40.8}  \\ 
        \midrule
        LLaMA-13B  & 16 & 40.6 & 36.7 & 48.9 & 48.0 & 43.3  \\
        \hdashline[0.5pt/2pt]
        NormalFloat & 4 & 43.0 & 34.5 & 51.8 & 51.4 & 45.0  \\
        PEQA &  4 & 43.0 & 37.7 & 53.6 & 49.0 & 45.0  \\ 
        QLoRA &  4 & 45.4 & 37.4 & 55.7 & 54.3 & 48.0  \\ 
        QLoRA {\scriptsize w/ GPTQ} & 4 & 48.4 & 38.3 & 54.9 & 55.2 & 49.2 \\
        QA-LoRA &  4 & 48.4 & 38.3 & 54.9 & 55.2 & 49.2  \\  
        \textbf{IR-QLoRA} &  4 & 47.2 & 39.0 & 56.5 & 55.0 & \textbf{49.3}  \\ 
        \midrule 
        LLaMA-30B  & 16 & 56.2 & 45.9 & 67.1 & 63.9 & 58.2  \\ 
        \hdashline[0.5pt/2pt]
        NormalFloat & 4 & 55.3 & 44.7 & 66.2 & 63.3 & 57.3  \\
        QLoRA &  4 & 55.4 & 46.0 & 66.4 & 63.6 & 57.7  \\ 
        QLoRA {\scriptsize w/ GPTQ} & 4 & 55.8 & 46.4 & 67.0 & 64.0 & 58.1 \\
        QA-LoRA &  4 & 55.8 & 46.4 & 67.0 & 64.0 & 58.1  \\ 
        \textbf{IR-QLoRA} &  4 & 56.7 & 46.7 & 66.5 & 63.2 & \textbf{58.2}  \\ 
        \midrule
        LLaMA-65B & 16 & 61.4 & 51.9 & 73.6 & 67.6 & 63.4  \\ 
        \hdashline[0.5pt/2pt]
        QA-LoRA &  4 & 60.8 & 50.5 & 72.5 & 66.7 & 62.5 \\ 
        NormalFloat & 4 & 60.7 & 52.3 & 72.6 & 67.3 & 63.0  \\
        QLoRA {\scriptsize w/ GPTQ} & 4 &  60.4 & 52.5 & 73.0 & 67.2 & 63.0 \\
        QLoRA &  4 & 60.3 & 52.7 & 72.9 & 67.4 & 63.1 \\  
        \textbf{IR-QLoRA} & 4 & 60.1 & 50.1 & 74.4 & 68.7 & \textbf{63.1}  \\ 
    \bottomrule
    \end{tabular}}
\end{table}

\subsection{Main Results}
To evaluate the performance of IR-QLoRA, we conducted comprehensive experiments and compared our IR-QLoRA with the state-of-the-art (SOTA) LoRA-finetuning quantization methods, \ie, QLoRA~\cite{dettmers2023qlora} and QA-LoRA~\cite{xu2023qalora}. We also compare with PEQA~\cite{kim2023memory} without LoRA follow~\cite{xu2023qalora}.
Table~\ref{tab:mmlu_alpaca} and Table~\ref{tab:mmlu_Flan v2} present the 5-shot accuracy results on the MMLU benchmark finetuned on the Alpaca~\cite{taori2023stanford} and Flan v2~\cite{longpre2023flan} datasets, respectively.

\begin{table}[t]
    \centering
    \caption{Accuracy (\%) comparison of LLaMA on the MMLU finetuned on the Flan v2 dataset}
    \vspace{-0.1in}
    \label{tab:mmlu_Flan v2}
    \setlength{\tabcolsep}{.7mm}
    {
    \begin{tabular}{lcccccc}
    \toprule
        \multirow{2}{*}{\textbf{Method}} & \multirow{2}{*}{\textbf{\#Bit}} & \multicolumn{5}{c}{\textbf{MMLU}} \\ \cmidrule(lr{.75em}){3-7} 
        ~ & ~ & \textbf{Hums.} & \textbf{STEM} & \textbf{Social} & \textbf{Other} & \textbf{Avg.}  \\ \midrule
        LLaMA-7B & 16 & 33.3 & 29.8 & 37.8 & 38.0 & 34.6  \\
        \hdashline[0.5pt/2pt]
        NormalFloat & 4 & 33.1 & 30.6 & 38.8 & 38.8 & 35.1 \\
        QLoRA {\scriptsize w/ GPTQ} & 4 & 33.8 & 31.3 & 37.4 & 42.2 & 36.0  \\
        QLoRA & 4 & 41.4 & 35.0 & 49.8 & 52.0 & 44.3  \\ 
        QA-LoRA & 4  & 43.9 & 38.0 & 54.3 & 53.0 & 47.0 \\ 
        \textbf{IR-QLoRA} & 4  & 44.2 & 39.3 & 54.5 & 52.9 & \textbf{47.4} \\ 
        \midrule
        LLaMA-13B & 16 & 40.6 & 36.7 & 48.9 & 48.0 & 43.3  \\
        \hdashline[0.5pt/2pt]
        NormalFloat & 4 & 43.0 & 34.5 & 51.8 & 51.4 & 45.0  \\
        QLoRA {\scriptsize w/ GPTQ} & 4 & 48.4 & 38.3 & 54.9 & 55.2 & 49.2 \\
        QLoRA &  4 & 49.9 & 40.1 & 60.2 & 57.9 & 51.9 \\  
        QA-LoRA &  4 & 50.0 & 41.5 & 60.5 & 58.4 & 52.4  \\ 
        \textbf{IR-QLoRA} &  4 & 49.2 & 41.2 & 62.1 & 59.2 & \textbf{52.6} \\ 
        \midrule
        LLaMA-30B & 16 & 56.2 & 45.9 & 67.1 & 63.9 & 58.2  \\ 
        \hdashline[0.5pt/2pt]
        NormalFloat & 4 & 55.3 & 44.7 & 66.2 & 63.3 & 57.3  \\
        QLoRA {\scriptsize w/ GPTQ} & 4 & 55.8 & 46.4 & 67.0 & 64.0 & 58.1 \\
        QLoRA &  4 & 57.2 & 48.6 & 69.8 & 65.2 & 60.0 \\ 
        QA-LoRA &  4 & 57.9 & 48.8 & 71.0 & 65.5 & 60.6  \\ 
        \textbf{IR-QLoRA} &  4 & 58.1 & 49.4 & 70.7 & 65.8 & \textbf{60.8}  \\ 
        \midrule
        LLaMA-65B & 16 & 61.4 & 51.9 & 73.6 & 67.6 & 63.4  \\ 
        \hdashline[0.5pt/2pt]
        NormalFloat & 4 & 60.7 & 52.3 & 72.6 & 67.3 & 63.0  \\
        QLoRA {\scriptsize w/ GPTQ} & 4 &  60.4 & 52.5 & 73.0 & 67.2 & 63.0 \\
        QLoRA &  4 & 59.8 & 52.9 & 75.0 & 69.6 & 63.9  \\  
        QA-LoRA &  4 & 57.6 & 51.1 & 73.9 & 67.4 & 62.1 \\ 
        \textbf{IR-QLoRA} &  4 & 61.6 & 52.0 & 75.6 & 68.9 & \textbf{64.3} \\ 
    \bottomrule
    \end{tabular}}
\end{table}

Comprehensive results indicate that across various sizes of LLaMA models, IR-QLoRA consistently outperforms all comparative quantization methods by a convincing margin. Compared to the baseline method QLoRA, our IR-QLoRA achieves a significant improvement in accuracy on the MMLU benchmark under the same finetuning pipeline. Specifically, as shown in Table~\ref{tab:mmlu_alpaca}, the 4-bit LLaMA-7B model finetuned with IR-QLoRA on the Alpaca dataset achieves an accuracy of 40.8\%, significantly surpassing the model obtained with QLoRA at 38.4\%. This outstanding trend continues in larger LLaMA-13B and LLaMA-30B models, where IR-QLoRA exceeds the baseline by 1.3\%, 0.5\%, respectively.
Compared to QLoRA with GPTQ and SOTA QA-LoRA using integer quantization, our method consistently performs better across various settings, with a notable advantage of 1.4\% even on LLaMA-7B. We further provide results for variants of the IR-QLoRA integer quantizer in Section~\ref{sec:Analysis_and_Discussion} below to demonstrate the robust improvement of our techniques in IR-QLoRA across different quantizers.

Table~\ref{tab:mmlu_Flan v2} presents the results obtained using Flan v2~\cite{longpre2023flan} as the finetuning dataset. Similar to the results on the Alpaca dataset, IR-QLoRA consistently achieves optimal results and outperforms SOTA methods across various settings, and even improves performance after finetuning with this dataset. For instance, the comparison of IR-QLoRA results on LLaMA-7B is Alpaca 47.4\% vs. Flan v2 40.8\%, and the average improvement of IR-QLoRA over QLoRA across various sizes of LLaMA is 1.25\%. This indicates that IR-QLoRA consistently provides stable benefits when using different finetuning datasets.

In addition, we conduct the accuracy comparison on the recently proposed LLaMA2, demonstrating the generalization performance of the proposed IR-QLoRA across LLM families.
Specifically, we applied IR-QLoRA to the 7B and 13B models of LLaMA2 and compared their evaluation results on the MMLU benchmark with the QA-LoRA method, which currently holds SOTA performance.
The results in Table~\ref{tab:mmlu_LLaMA2} show that our method not only achieved 2.7\% performance improvement but also demonstrated advantages in almost every individual metric on LLaMA2-7B. These results indicate that IR-QLoRA exhibits strong generalization across different LLM families.

\begin{table}[t]
    \captionsetup{justification=centering,singlelinecheck=false}
    \centering
    \caption{Accuracy (\%) comparison of LLaMA2 on MMLU}
    \vspace{-0.1in}
    \label{tab:mmlu_LLaMA2}
    \setlength{\tabcolsep}{0.1mm}
    {
    \begin{tabular}{lcccccccc}
    \toprule
        \multirow{2}{*}{\textbf{Method}} & \multirow{2}{*}{\textbf{Dataset}} & \multirow{2}{*}{\textbf{\#Bit}} & \multicolumn{5}{c}{\textbf{MMLU}} \\ \cmidrule(lr{.75em}){4-8} 
        ~ & ~ & ~ & \textbf{Hums.} & \textbf{STEM} & \textbf{Social} & \textbf{Other} & \textbf{Avg.}  \\ \midrule
        LLaMA2-7B & - & 16 & 43.0 & 36.4 & 51.4 & 52.2 & 45.5 \\
        \hdashline[0.5pt/2pt]
        NormalFloat & - & 4 & 42.0 & 35.9 & 51.0 & 51.4 & 44.8 \\
        \hdashline[0.5pt/2pt]
        QA-LoRA & Alpaca & 4 & 42.1 & 34.4 & 49.1 & 50.3 & 43.9 \\
        \textbf{IR-QLoRA} & Alpaca & 4 & 43.4 & 36.8 & 51.9 & 53.6 & \textbf{46.2} \\ 
        \hdashline[0.5pt/2pt]
        QA-LoRA & Flan v2 & 4 & 48.4 & 41.4 & 59.4 & 58.6 & 51.7 \\
        \textbf{IR-QLoRA} & Flan v2 & 4 & 49.2 & 41.6 & 60.2 & 58.0 & \textbf{52.0} \\ 
        \midrule
        LLaMA2-13B & - & 16 & 53.3 & 44.1 & 63.3 & 61.0 & 55.3 \\
        \hdashline[0.5pt/2pt]
        NormalFloat & - & 4 & 52.2 & 44.1 & 62.3 & 60.8 & 54.7 \\
        \hdashline[0.5pt/2pt]
        QA-LoRA & Alpaca & 4 & 48.0 & 43.0 & 59.7 & 57.4 & 51.7 \\
        \textbf{IR-QLoRA} & Alpaca & 4 & 51.9 & 43.9 & 61.9 & 60.4 & \textbf{54.4} \\ 
        \hdashline[0.5pt/2pt]
        QA-LoRA & Flan v2 & 4 & 52.9 & 44.8 & 65.9 & 64.0 & \textbf{56.6}  \\
        \textbf{IR-QLoRA} & Flan v2 & 4 & 53.1 & 45.6 & 64.9 & 63.8 & 56.5 \\ 
    \bottomrule
    \end{tabular}}
\end{table}

\subsection{Ablation Study}
To reveal the effectiveness of techniques in the proposed IR-QLoRA on accuracy and efficiency, we conduct extensive ablation studies for 4-bit LLaMA-7B on MMLU.

\begin{figure}[tb]
\centering
\includegraphics[width=1.\linewidth]{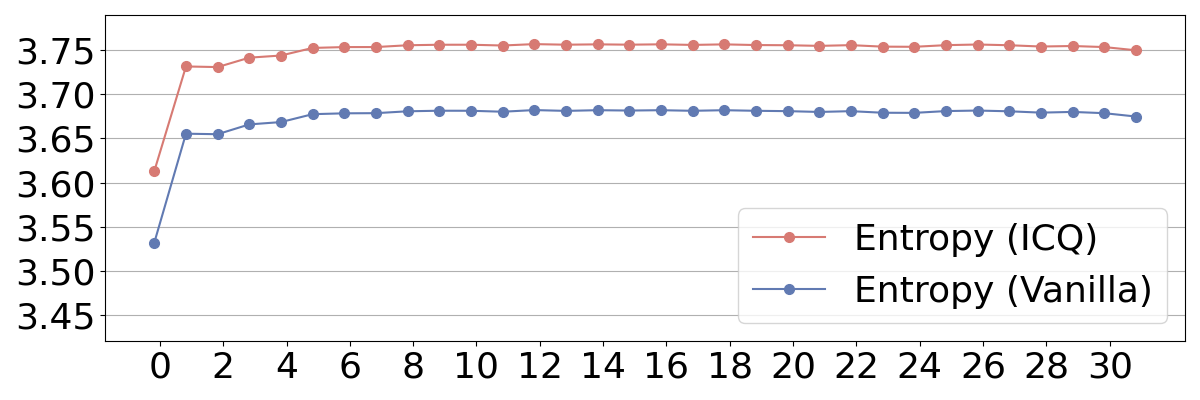}
\caption{Entropy of linear projections in LLaMA-7B}
\label{fig:Entopy_key_proj}
\end{figure}

\textbf{Accuracy Ablation.}
Figure~\ref{fig:Entopy_key_proj} illustrates the entropy of quantized weights with ICQ is constantly higher than the vanilla QLoRA across different layers. This indicates that ICQ effectively enhances the mutual information between quantized weights of LLMs and original counterparts, thereby reducing information loss and producing a 1.9\% accuracy gain, as Table~\ref{tab:ablation} shows.
We also conducted different controlled experiments to illustrate the effectiveness of the IEC. Firstly, we apply the IEC directly to Vanilla, resulting in 1.8\% enhancements, as indicated by the results in Table~\ref{tab:ablation}. Then we conduct in-depth observations, applying IEC on the first and latter sub-units respectively, both of which lead to improvements, (IEC {\scriptsize ($U_1$)} 1.0\% \textit{vs.} IEC {\scriptsize ($U_2$)} 1.3\%). Moreover, when ICQ and IEC are combined, their synergistic effect surpasses individual contributions, pushing the quantized model to achieve up to 40.8\% accuracy.

We also present the performances of the quantized LLMs solely with our proposed ICQ in IR-QLoRA without LoRA and fine-tuning, including the detailed accuracy in the MMLU benchmark and the average information entropy of quantized LLMs. 
As Table~\ref{tab:ablation_ent} shows, ICQ can solely improve the entropy of quantized weight in LLMs even without finetuning while contributing significantly to the experiment results compared to NormalFloat quantization (average accuracy increases by 0.5\%). We also compare the average information entropy between quantized LLM with and without our ICQ. With the proposed ICQ of IR-QLoRA, the average entropy increases by 0.07, which means a significant improvement of the information retention for quantized weights in LLMs, proving the accuracy improvement of ICQ comes from its motivation of information retention.

\begin{table}[t]
    \centering
    \caption{Accuracy (\%) ablation on MMLU}
    \vspace{-0.1in}
    \label{tab:ablation}
    \setlength{\tabcolsep}{1.1mm}
    {
    \begin{threeparttable}
    \begin{tabular}{lcccccc}
    \toprule
        \multirow{2}{*}{\textbf{Method}} & \multirow{2}{*}{\textbf{\#Bit}} & \multicolumn{5}{c}{\textbf{MMLU}} \\ \cmidrule(lr{.75em}){3-7} 
        ~ & ~ & \textbf{Hums.} & \textbf{STEM} & \textbf{Social} & \textbf{Other} & \textbf{Avg.}  \\ \midrule
        LLaMA-7B & 16 & 33.3 & 29.8 & 37.8 & 38.0 & 34.6  \\
        \hdashline[0.5pt/2pt]
        Vanilla & 4 & 36.1 & 31.9 & 42.0 & 44.5 & 38.4  \\ 
        ICQ & 4 & 37.9 & 33.6 & 43.9 & 46.7 & 40.3  \\ 
        IEC {\scriptsize ($U_1$)} & 4 & 37.9 & 31.9 & 43.4 & 44.8 & 39.4  \\ 
        IEC {\scriptsize ($U_2$)} & 4 & 38.0 & 32.3 & 43.6 & 45.1 & 39.7  \\ 
        IEC & 4 & 38.3 & 33.0 & 44.5 & 45.7 & 40.2  \\ 
        \textbf{IR-QLoRA} & 4 & 38.6 & 34.6 & 45.2 & 45.5 & \textbf{40.8}  \\ 
    \bottomrule
    \end{tabular}
    \begin{tablenotes}
    \footnotesize
    \item[1] ICQ and IEC denote the vanilla QLoRA with ICQ and IEC, and IR-QLoRA uses both of them. IEC {\scriptsize ($U_1$)} and IEC {\scriptsize ($U_2$)} denote the further ablation for IEC in the first or latter LoRA sub-units
    \end{tablenotes}
    \end{threeparttable}
    }
\end{table}

\begin{table}[t]
    \centering
    \caption{Accuracy (\%) ablation on MMLU for ICQ without LoRA and fine-tuning}
    \vspace{-0.1in}
    \label{tab:ablation_ent}
    \setlength{\tabcolsep}{.55mm}
    {
    \begin{threeparttable}
    \begin{tabular}{lccccccc}
    \toprule
    \textbf{Method} & \textbf{\#Bit} & \textbf{Hums.} & \textbf{STEM} & \textbf{Social} & \textbf{Other} & \textbf{Avg.} & \textbf{Ent.} \\ 
    \midrule
    LLaMA-7B & 16 & 33.3 & 29.8 & 37.8 & 38.0 & 34.6 & / \\
    \hdashline[0.5pt/2pt]
    NormalFloat & 4 & 33.1 & 30.6 & 38.8 & 38.8 & 35.1 & 3.67 \\
    ICQ & 4 & 33.6 & 31.7 & 39.6 & 38.2 & \textbf{35.6} & \textbf{3.74} \\
    \bottomrule
    \end{tabular}
    \end{threeparttable}
    }
\end{table}

\begin{table}[t]
    \centering
    \caption{Efficiency ablation on the different sizes of LLaMA}
    \vspace{-0.1in}
    \label{tab:ablation_params}
    \setlength{\tabcolsep}{2.35mm}
    {
    \begin{tabular}{lc rrrr}
    \toprule
        \multirow{2}{*}{\textbf{Method}} & \multirow{2}{*}{\textbf{\#Bit}} & \multicolumn{2}{c}{\textbf{\textbf{\#Params}{\scriptsize (GB)}}} & \multicolumn{2}{c}{\textbf{Time}{\scriptsize (h)}} \\ \cmidrule(lr{.75em}){3-6} 
        ~ & ~ & \textbf{7B} & \textbf{13B}  & \textbf{7B} & \textbf{13B} \\
        \midrule
        LLaMA & 16 & 12.55 & 24.24 & - & - \\
        \hdashline[0.5pt/2pt]
        Vanilla & 4 & 2.34 & 4.13 & 15.33 & 26.18  \\ 
        ICQ & 4 & 2.39 & 4.22 & 15.40 & 26.26 \\ 
        IEC & 4 & 2.34 & 4.13 & 15.33 & 26.18  \\ 
        \textbf{IR-QLoRA}  & 4 & 2.39 & 4.22 & 15.40 & 26.26 \\  
    \bottomrule
    \end{tabular}}
\end{table}

\begin{table*}[t]
    \centering
    \caption{Additional fine-tuning time (h) for different sizes of LLaMA}
    \vspace{-0.1in}
    \label{tab:ablation_eff_train}
    \setlength{\tabcolsep}{5.2mm}
    {
    \begin{threeparttable}
    \begin{tabular}{lrrrr}
    \toprule
    \textbf{Model} & \textbf{LLaMA-7B} & \textbf{LLaMA-13B} & \textbf{LLaMA-30B} & \textbf{LLaMA-65B} \\ \midrule
    Original Training Time (h) & 15.40 & 26.26 & 43.07 & 119.38 \\ 
    Additional Training Time (h) & 0.07 & 0.08 & 0.36 & 0.41 \\ 
    \bottomrule
    \end{tabular}
    \end{threeparttable}
    }
    \vspace{-0.05in}
\end{table*}

\begin{table*}[htb]
    \centering
    \caption{Accuracy (\%) comparison on the Commonsense QA datasets}
    \vspace{-0.1in}
    \label{tab:commonsenseqa}
    \setlength{\tabcolsep}{2.55mm}
    {
    \begin{tabular}{lccccccccc}
    \toprule
        \multirow{2}{*}{\textbf{Method}} & \multirow{2}{*}{\textbf{\#Bit}} & \multicolumn{8}{c}{\textbf{CommonsenseQA}} \\ 
        \cmidrule(lr{.75em}){3-10} 
        ~ & ~ & \textbf{HellaSwag} & \textbf{PIQA} & \textbf{WinoGrande} & \textbf{ARC-e} & \textbf{ARC-c} & \textbf{BoolQ} & \textbf{OBQA} & \textbf{Avg.} \\ \midrule
        LLaMA-7B & 16 & 56.3 & 78.2 & 67.1 & 67.3 & 38.2 & 72.9 & 28.4 & 58.3   \\
        \hdashline[0.5pt/2pt]
        NormalFloat & 4 & 56.7 & 78.7 & 70.6 & 75.7 & 41.6 & 74.7 & 33.2 & 61.6   \\ 
        QLoRA {\scriptsize w/ GPTQ} & 4 & 57.4 & 77.6 & 66.2 & 70.9 & 41.8 & 73.5 & 31.2 & 59.8 \\
        QLoRA & 4 &  61.8 & 78.1 & 68.4 & 75.8 & 43.6 & 73.7 & 32.8 & 62.0  \\ 
        QA-LoRA & 4 & 58.6 & 78.0 & 66.9 & 71.2 & 43.9 & 79.9 & 34.0 & 61.8  \\ 
        \textbf{IR-QLoRA} & 4 & 54.7 & 78.8 & 72.6 & 76.6 & 45.1 & 80.6 & 37.2 & \textbf{63.7}  \\ 
    \bottomrule
    \end{tabular}}
    \vspace{-0.05in}
\end{table*}

\textbf{Efficiency Ablation.}
Table~\ref{tab:ablation_params} demonstrates that the proposed ICQ and IEC techniques impose little additional storage and training overhead. For ICQ, the added parameters are only equivalent to the quantized scaling factor, and double quantization is applied to reduce storage further. Therefore, the additional storage introduced by ICQ is minor, increasing only by 2.04\% on the 4-bit LLaMA-7B. The optimization process for $\tau$ also adds only a negligible amount of training time (\eg 0.46\% for LLamA-7B and 0.31\% for LLaMA-13B). Furthermore, this additional time is exclusively required for the initial optimization during training and does not result in increased inference time. IEC just introduces 2 additional parameters per layer, which can be negligible in the whole model. In the case of IR-QLoRA, our ICQ and IEC significantly enhance the accuracy performance of quantized LLMs with little additional storage increase. The ablation studies present the effectiveness and efficiency of ICQ and IEC, showcasing their strong capabilities in constructing accurate and efficient LLMs.

As for the fine-tuning efficiency, the additional time incurred by IEC can be considered negligible theoretically, whereas the increased fine-tuning time due to ICQ depends on the range set during the search process. The larger the range and the finer the granularity, the longer the search process will take. We represented the additional fine-tuning time using the default setting ($\lambda=0.1, n=100$) in Table~\ref{tab:ablation_eff_train}, where the original fine-tuning time is the time used for baseline fine-tuning, and the additional fine-tuning time is the additional time used by IR-QLoRA fine-tuning. Compared with the original fine-tuning time, the additional fine-tuning time is only up to 0.84\%, which is extremely low additional overhead.

\subsection{Analysis and Discussion}
\label{sec:Analysis_and_Discussion}

\textbf{IR-QLoRA on More Evaluation Benchmark}.
We present the 0-shot results of the CommonsenseQA benchmark in Table~\ref{tab:commonsenseqa}. Similar to the phenomenon on the MMLU benchmark, our IR-QLoRA consistently maintains the best average accuracy for LLaMA-7B on the CommonsenseQA benchmark compared to SOTA methods, and also significantly improves the effectiveness in the majority of sub-items. More evaluation results are presented in Appendix~\ref{sec:Case_Study}.

\begin{table}[!h]
    \centering
    \vspace{-0.1in}
    \caption{Accuracy (\%) comparison under 2-3 bits on MMLU}
    \vspace{-0.1in}
    \label{tab:nf2nf3}
    \setlength{\tabcolsep}{0.05mm}
    {
    \begin{tabular}{lccccccc}
    \toprule
        \multirow{2}{*}{\textbf{Method}} & \multirow{2}{*}{\textbf{Data}} & \multirow{2}{*}{\textbf{\#Bit}} & \multicolumn{5}{c}{\textbf{MMLU}} \\ \cmidrule(lr{.75em}){4-8} 
        ~ & ~ & ~ & \textbf{Hums.} & \textbf{STEM} & \textbf{Social} & \textbf{Other} & \textbf{Avg.}  \\ \midrule
        LLaMA-7B & - & 16 & 33.3 & 29.8 & 37.8 & 38.0 & 34.6  \\
        \midrule
        NormalFloat & - & 3 & 30.5 & 29.9 & 34.8 & 34.9 & 32.3 \\
        \hdashline[0.5pt/2pt]
        QLoRA {\scriptsize w/ GPTQ} & Alpaca & 3 & 31.6 & 30.1 & 35.6 & 39.8 & 34.0 \\
        QLoRA & Alpaca & 3 & 35.8 & 32.1 & 40.7 & 43.1 & 37.8 \\
        QA-LoRA & Alpaca & 3 & 35.6 & 30.5 & 41.5 & 42.7 & 37.4  \\ 
        \textbf{IR-QLoRA} & Alpaca & 3 & 36.0 & 33.9 & 42.2 & 42.7 & \textbf{38.4} \\ 
        \hdashline[0.5pt/2pt]
        QLoRA {\scriptsize w/ GPTQ} & Flan v2 & 3 & 32.2 & 31.7 & 42.7 & 42.8 & 36.9 \\
        QLoRA & Flan v2 & 3 & 41.3 & 37.1 & 50.9 & 49.8 & 44.5 \\ 
        QA-LoRA & Flan v2 & 3 & 41.3 & 36.0 & 52.8 & 50.2 & 44.7  \\ 
        \textbf{IR-QLoRA} & Flan v2 & 3 &  43.0 & 37.7 & 52.3 & 51.7 & \textbf{45.9}  \\ 
        \midrule
        NormalFloat & - & 2 & 24.2 & 28.9 & 31.1 & 25.0 & 26.9 \\
        \hdashline[0.5pt/2pt]
        QLoRA {\scriptsize w/ GPTQ} & Alpaca & 2 & 23.4 & 26.2 & 26.4 & 28.4 & 25.8 \\
        QLoRA & Alpaca & 2 & 24.0 & 27.0 & 27.5 & 26.7 & 26.2 \\ 
        QA-LoRA & Alpaca & 2 & 27.3 & 26.1 & 26.1 & 30.3 & 27.5 \\ 
        \textbf{IR-QLoRA} & Alpaca & 2 & 26.0 & 27.8 & 30.2 & 28.3 & \textbf{27.8}  \\
        \hdashline[0.5pt/2pt]
        QLoRA {\scriptsize w/ GPTQ} & Flan v2 & 2 & 23.9 & 25.3 & 26.2 & 25.3 & 25.0 \\
        QLoRA & Flan v2 & 2 & 31.8 & 28.7 & 36.7 & 37.7 & 33.5\\ 
        QA-LoRA & Flan v2 & 2 & 31.8 & 28.1 & 34.5 & 38.5 & 33.2  \\ 
        \textbf{IR-QLoRA} & Flan v2 & 2 & 31.7 & 29.4 & 37.8 & 36.5 & \textbf{33.7}  \\
    \bottomrule
    \end{tabular}}
    \vspace{-0.1in}
\end{table}

\textbf{IR-QLoRA under Ultra-low Bit-width}.
We have evaluated and compared the proposed IR-QLoRA under ultra-low bit-width. Specifically, we employed the quantization methods from QLoRA~\cite{dettmers2023qlora} and LoftQ~\cite{li2023loftq}, following the percentile quantization approach to construct NF2 and NF3 quantization. Additionally, we adhered to the 2-bit and 3-bit integer quantization results from QA-LoRA and QLoRA with GPTQ as presented in~\cite{xu2023qalora}. Table~\ref{tab:nf2nf3} demonstrates that as the quantization bit-width decreases, the performance of the baseline QLoRA sharply declines, to the extent that its performance in the 2-bit scenario is similar to random. In contrast, our IR-QLoRA exhibits superior performance, with only a 0.9\% accuracy difference compared to the 16-bit counterpart when finetuning a 2-bit model on the Flan v2 dataset. These results strongly indicate the competitiveness of our IR-QLoRA in the realm of ultra-low bit-width. The quantization values for NF quantization are presented in Appendix~\ref{sec:NormalFloat_Quantization}.

\textbf{IR-QLoRA with Integer Quantizer}.
We demonstrate the strong generality of the technology in IR-QLoRA across different quantization frameworks and present variants based on QA-LoRA as the baseline. In this variant, ICQ performs searching for the zero point and determines it along with the scaling factor for integer quantizers, while the universality of IEC for LoRA allows direct application for the QA-LoRA baseline. As shown in Table~\ref{tab:QALoRA_ours}, experiments reveal a significant improvement in accuracy for IR-QLoRA under the same quantizer form and neural architecture. Furthermore, we emphasize that due to the zero point carried by integer quantizers themselves, where the calibration constant $\tau_\text{ICQ}$ in our ICQ can be merged, the improvements brought about by our techniques come at almost zero cost. These experiments indicate that the techniques in our QA-LoRA can be effectively integrated into various LoRA-finetuning quantization methods for LLMs and bring general benefits.

\begin{table}[tb]
    \centering
    \vspace{-0.1in}
    \caption{Comparsion of IR-QLoRA variants on MMLU}
    \vspace{-0.1in}
    \label{tab:QALoRA_ours}
    \setlength{\tabcolsep}{0.1mm}
    {
    \begin{tabular}{lccccccc}
    \toprule
        \multirow{2}{*}{\textbf{Method}} & \multirow{2}{*}{\textbf{\#Bit}} & \multicolumn{5}{c}{\textbf{MMLU}} \\ \cmidrule(lr{.75em}){3-8} 
        ~ & ~ & \textbf{Hums.} & \textbf{STEM} & \textbf{Social} & \textbf{Other} & \textbf{Avg.}  \\ \midrule
        LLaMA-7B & 16 & 33.3 & 29.8 & 37.8 & 38.0 & 34.6  \\
        \hdashline[0.5pt/2pt]
        QA-LoRA & 4 & 36.6 & 32.4 & 44.8 & 44.9 & 39.4  \\ 
        \textbf{IR-QLoRA {\scriptsize (QA-LoRA)}} & 4 & 37.3 & 32.8 & 43.8 & 46.7 & \textbf{39.9} \\
    \bottomrule
    \end{tabular}}
    \vspace{-0.2in}
\end{table}

\section{Conclusion}
This paper introduces the \textbf{IR-QLoRA}, designed to accurately quantize LLMs with LoRA-finetuning via information retention. This framework leverages two key technologies: statistics-based \textit{Information Calibration Quantization}, which ensures that the quantized parameters of the LLM accurately retain the original information; and finetuning-based \textit{Information Elastic Connection}, enabling LoRA to employ elastic representation transformation with diverse information. 
Extensive experiments validate that IR-QLoRA delivers convincing accuracy improvements across the LLaMA and LLaMA2 families, even with 2-4 bit-widths, accompanied by a minimal 0.45\% increase in time consumption. Remarkably versatile, IR-QLoRA seamlessly integrates with various quantization frameworks.
In a nutshell, our IR-QLoRA significantly advances the accuracy of LoRA-finetuning quantization for LLMs, facilitating practical deployment in resource-constrained scenarios.

\textbf{Acknowledgement.}
This work was supported by the National Science and Technology Major Project (2022ZD0116405), the Swiss National Science Foundation (SNSF) project 200021E\_219943 Neuromorphic Attention Models for Event Data (NAMED), the Baidu Scholarship, and the National Natural Science Foundation of China (No. 62306025, No. 92367204, No. 62276014).

\section*{Impact Statement}

This paper presents work whose goal is to advance the field of Machine Learning. There are many potential societal consequences of our work, none which we feel must be specifically highlighted here.

{
\bibliography{CR/IR-QLoRA}

\begin{thebibliography}{48}
\providecommand{\natexlab}[1]{#1}
\providecommand{\url}[1]{\texttt{#1}}
\expandafter\ifx\csname urlstyle\endcsname\relax
  \providecommand{\doi}[1]{doi: #1}\else
  \providecommand{\doi}{doi: \begingroup \urlstyle{rm}\Url}\fi

\bibitem[Agarwal et~al.(2023)Agarwal, Vieillard, Stanczyk, Ramos, Geist, and
  Bachem]{agarwal2023gkd}
Agarwal, R., Vieillard, N., Stanczyk, P., Ramos, S., Geist, M., and Bachem, O.
\newblock Gkd: Generalized knowledge distillation for auto-regressive sequence
  models.
\newblock \emph{arXiv preprint arXiv:2306.13649}, 2023.

\bibitem[Baskin et~al.(2021)Baskin, Liss, Schwartz, Zheltonozhskii, Giryes,
  Bronstein, and Mendelson]{baskin2021uniq}
Baskin, C., Liss, N., Schwartz, E., Zheltonozhskii, E., Giryes, R., Bronstein,
  A.~M., and Mendelson, A.
\newblock Uniq: Uniform noise injection for non-uniform quantization of neural
  networks.
\newblock \emph{ACM Transactions on Computer Systems (TOCS)}, 37\penalty0
  (1-4):\penalty0 1--15, 2021.

\bibitem[Bisk et~al.(2020)Bisk, Zellers, Gao, Choi, et~al.]{bisk2020piqa}
Bisk, Y., Zellers, R., Gao, J., Choi, Y., et~al.
\newblock Piqa: Reasoning about physical commonsense in natural language.
\newblock In \emph{Proceedings of the AAAI conference on artificial
  intelligence}, volume~34, pp.\  7432--7439, 2020.

\bibitem[Brown et~al.(2020)Brown, Mann, Ryder, Subbiah, Kaplan, Dhariwal,
  Neelakantan, Shyam, Sastry, Askell, et~al.]{brown2020language}
Brown, T., Mann, B., Ryder, N., Subbiah, M., Kaplan, J.~D., Dhariwal, P.,
  Neelakantan, A., Shyam, P., Sastry, G., Askell, A., et~al.
\newblock Language models are few-shot learners.
\newblock \emph{Advances in neural information processing systems},
  33:\penalty0 1877--1901, 2020.

\bibitem[Chang et~al.(2023)Chang, Wang, Wang, Wu, Zhu, Chen, Yang, Yi, Wang,
  Wang, et~al.]{chang2023survey}
Chang, Y., Wang, X., Wang, J., Wu, Y., Zhu, K., Chen, H., Yang, L., Yi, X.,
  Wang, C., Wang, Y., et~al.
\newblock A survey on evaluation of large language models.
\newblock \emph{arXiv preprint arXiv:2307.03109}, 2023.

\bibitem[Chitty-Venkata et~al.(2023)Chitty-Venkata, Mittal, Emani, Vishwanath,
  and Somani]{chitty2023survey}
Chitty-Venkata, K.~T., Mittal, S., Emani, M., Vishwanath, V., and Somani, A.~K.
\newblock A survey of techniques for optimizing transformer inference.
\newblock \emph{Journal of Systems Architecture}, pp.\  102990, 2023.

\bibitem[Chowdhery et~al.(2023)Chowdhery, Narang, Devlin, Bosma, Mishra,
  Roberts, Barham, Chung, Sutton, Gehrmann, et~al.]{chowdhery2023palm}
Chowdhery, A., Narang, S., Devlin, J., Bosma, M., Mishra, G., Roberts, A.,
  Barham, P., Chung, H.~W., Sutton, C., Gehrmann, S., et~al.
\newblock Palm: Scaling language modeling with pathways.
\newblock \emph{Journal of Machine Learning Research}, 24\penalty0
  (240):\penalty0 1--113, 2023.

\bibitem[Clark et~al.(2019)Clark, Lee, Chang, Kwiatkowski, Collins, and
  Toutanova]{clark2019boolq}
Clark, C., Lee, K., Chang, M.-W., Kwiatkowski, T., Collins, M., and Toutanova,
  K.
\newblock Boolq: Exploring the surprising difficulty of natural yes/no
  questions.
\newblock \emph{arXiv preprint arXiv:1905.10044}, 2019.

\bibitem[Clark et~al.(2018)Clark, Cowhey, Etzioni, Khot, Sabharwal, Schoenick,
  and Tafjord]{clark2018think}
Clark, P., Cowhey, I., Etzioni, O., Khot, T., Sabharwal, A., Schoenick, C., and
  Tafjord, O.
\newblock Think you have solved question answering? try arc, the ai2 reasoning
  challenge.
\newblock \emph{arXiv preprint arXiv:1803.05457}, 2018.

\bibitem[Dettmers et~al.(2021)Dettmers, Lewis, Shleifer, and
  Zettlemoyer]{dettmers20218}
Dettmers, T., Lewis, M., Shleifer, S., and Zettlemoyer, L.
\newblock 8-bit optimizers via block-wise quantization.
\newblock \emph{arXiv preprint arXiv:2110.02861}, 2021.

\bibitem[Dettmers et~al.(2022)Dettmers, Lewis, Belkada, and
  Zettlemoyer]{dettmers2022llm}
Dettmers, T., Lewis, M., Belkada, Y., and Zettlemoyer, L.
\newblock Llm. int8 (): 8-bit matrix multiplication for transformers at scale.
\newblock \emph{arXiv preprint arXiv:2208.07339}, 2022.

\bibitem[Dettmers et~al.(2023)Dettmers, Pagnoni, Holtzman, and
  Zettlemoyer]{dettmers2023qlora}
Dettmers, T., Pagnoni, A., Holtzman, A., and Zettlemoyer, L.
\newblock Qlora: Efficient finetuning of quantized llms.
\newblock \emph{arXiv preprint arXiv:2305.14314}, 2023.

\bibitem[Devlin et~al.(2018)Devlin, Chang, Lee, and Toutanova]{devlin2018bert}
Devlin, J., Chang, M.-W., Lee, K., and Toutanova, K.
\newblock Bert: Pre-training of deep bidirectional transformers for language
  understanding.
\newblock \emph{arXiv preprint arXiv:1810.04805}, 2018.

\bibitem[Frantar \& Alistarh(2023)Frantar and Alistarh]{frantar2023sparsegpt}
Frantar, E. and Alistarh, D.
\newblock Sparsegpt: Massive language models can be accurately pruned in
  one-shot.
\newblock In \emph{International Conference on Machine Learning}, pp.\
  10323--10337. PMLR, 2023.

\bibitem[Frantar et~al.(2022)Frantar, Ashkboos, Hoefler, and
  Alistarh]{frantar2022gptq}
Frantar, E., Ashkboos, S., Hoefler, T., and Alistarh, D.
\newblock Gptq: Accurate post-training quantization for generative pre-trained
  transformers.
\newblock \emph{arXiv preprint arXiv:2210.17323}, 2022.

\bibitem[Ganesh et~al.(2021)Ganesh, Chen, Lou, Khan, Yang, Sajjad, Nakov, Chen,
  and Winslett]{ganesh2021compressing}
Ganesh, P., Chen, Y., Lou, X., Khan, M.~A., Yang, Y., Sajjad, H., Nakov, P.,
  Chen, D., and Winslett, M.
\newblock Compressing large-scale transformer-based models: A case study on
  bert.
\newblock \emph{Transactions of the Association for Computational Linguistics},
  9:\penalty0 1061--1080, 2021.

\bibitem[Gao et~al.(2023)Gao, Tow, Abbasi, Biderman, Black, DiPofi, Foster,
  Golding, Hsu, Le~Noac'h, Li, McDonell, Muennighoff, Ociepa, Phang, Reynolds,
  Schoelkopf, Skowron, Sutawika, Tang, Thite, Wang, Wang, and
  Zou]{eval-harness}
Gao, L., Tow, J., Abbasi, B., Biderman, S., Black, S., DiPofi, A., Foster, C.,
  Golding, L., Hsu, J., Le~Noac'h, A., Li, H., McDonell, K., Muennighoff, N.,
  Ociepa, C., Phang, J., Reynolds, L., Schoelkopf, H., Skowron, A., Sutawika,
  L., Tang, E., Thite, A., Wang, B., Wang, K., and Zou, A.
\newblock A framework for few-shot language model evaluation, 12 2023.
\newblock URL \url{https://zenodo.org/records/10256836}.

\bibitem[Gu et~al.(2023)Gu, Dong, Wei, and Huang]{gu2023knowledge}
Gu, Y., Dong, L., Wei, F., and Huang, M.
\newblock Knowledge distillation of large language models.
\newblock \emph{arXiv preprint arXiv:2306.08543}, 2023.

\bibitem[Hendrycks et~al.(2020)Hendrycks, Burns, Basart, Zou, Mazeika, Song,
  and Steinhardt]{hendrycks2020measuring}
Hendrycks, D., Burns, C., Basart, S., Zou, A., Mazeika, M., Song, D., and
  Steinhardt, J.
\newblock Measuring massive multitask language understanding.
\newblock In \emph{International Conference on Learning Representations}, 2020.

\bibitem[Hendrycks et~al.(2021)Hendrycks, Burns, Basart, Zou, Mazeika, Song,
  and Steinhardt]{hendryckstest2021}
Hendrycks, D., Burns, C., Basart, S., Zou, A., Mazeika, M., Song, D., and
  Steinhardt, J.
\newblock Measuring massive multitask language understanding.
\newblock \emph{Proceedings of the International Conference on Learning
  Representations (ICLR)}, 2021.

\bibitem[Hu et~al.(2021)Hu, Shen, Wallis, Allen-Zhu, Li, Wang, Wang, and
  Chen]{hu2021lora}
Hu, E.~J., Shen, Y., Wallis, P., Allen-Zhu, Z., Li, Y., Wang, S., Wang, L., and
  Chen, W.
\newblock Lora: Low-rank adaptation of large language models.
\newblock \emph{arXiv preprint arXiv:2106.09685}, 2021.

\bibitem[Huang \& Chang(2022)Huang and Chang]{huang2022towards}
Huang, J. and Chang, K. C.-C.
\newblock Towards reasoning in large language models: A survey.
\newblock \emph{arXiv preprint arXiv:2212.10403}, 2022.

\bibitem[Huang et~al.(2022)Huang, Chen, Yu, and McKeown]{huang2022context}
Huang, Y., Chen, Y., Yu, Z., and McKeown, K.
\newblock In-context learning distillation: Transferring few-shot learning
  ability of pre-trained language models.
\newblock \emph{arXiv preprint arXiv:2212.10670}, 2022.

\bibitem[Jiang et~al.(2023)Jiang, Chan, Chen, and Wang]{jiang2023lion}
Jiang, Y., Chan, C., Chen, M., and Wang, W.
\newblock Lion: Adversarial distillation of closed-source large language model.
\newblock \emph{arXiv preprint arXiv:2305.12870}, 2023.

\bibitem[Kim et~al.(2023)Kim, Lee, Kim, Park, Yoo, Kwon, and
  Lee]{kim2023memory}
Kim, J., Lee, J.~H., Kim, S., Park, J., Yoo, K.~M., Kwon, S.~J., and Lee, D.
\newblock Memory-efficient fine-tuning of compressed large language models via
  sub-4-bit integer quantization.
\newblock \emph{arXiv preprint arXiv:2305.14152}, 2023.

\bibitem[Lee et~al.(2023)Lee, Jin, Kim, Kim, and Park]{lee2023owq}
Lee, C., Jin, J., Kim, T., Kim, H., and Park, E.
\newblock Owq: Lessons learned from activation outliers for weight quantization
  in large language models.
\newblock \emph{arXiv preprint arXiv:2306.02272}, 2023.

\bibitem[Li et~al.(2022)Li, Chen, Shen, Chen, Zhang, Li, Wang, Qian, Peng, Mao,
  et~al.]{li2022explanations}
Li, S., Chen, J., Shen, Y., Chen, Z., Zhang, X., Li, Z., Wang, H., Qian, J.,
  Peng, B., Mao, Y., et~al.
\newblock Explanations from large language models make small reasoners better.
\newblock \emph{arXiv preprint arXiv:2210.06726}, 2022.

\bibitem[Li et~al.(2023)Li, Yu, Liang, He, Karampatziakis, Chen, and
  Zhao]{li2023loftq}
Li, Y., Yu, Y., Liang, C., He, P., Karampatziakis, N., Chen, W., and Zhao, T.
\newblock Loftq: Lora-fine-tuning-aware quantization for large language models.
\newblock \emph{arXiv preprint arXiv:2310.08659}, 2023.

\bibitem[Lin et~al.(2023)Lin, Tang, Tang, Yang, Dang, and Han]{lin2023awq}
Lin, J., Tang, J., Tang, H., Yang, S., Dang, X., and Han, S.
\newblock Awq: Activation-aware weight quantization for llm compression and
  acceleration.
\newblock \emph{arXiv preprint arXiv:2306.00978}, 2023.

\bibitem[Liu et~al.(2023{\natexlab{a}})Liu, Liu, Gao, Gao, Zhao, Li, Ding, and
  Wen]{liu2023emergent}
Liu, P., Liu, Z., Gao, Z.-F., Gao, D., Zhao, W.~X., Li, Y., Ding, B., and Wen,
  J.-R.
\newblock Do emergent abilities exist in quantized large language models: An
  empirical study.
\newblock \emph{arXiv preprint arXiv:2307.08072}, 2023{\natexlab{a}}.

\bibitem[Liu et~al.(2023{\natexlab{b}})Liu, Oguz, Zhao, Chang, Stock, Mehdad,
  Shi, Krishnamoorthi, and Chandra]{liu2023llm}
Liu, Z., Oguz, B., Zhao, C., Chang, E., Stock, P., Mehdad, Y., Shi, Y.,
  Krishnamoorthi, R., and Chandra, V.
\newblock Llm-qat: Data-free quantization aware training for large language
  models.
\newblock \emph{arXiv preprint arXiv:2305.17888}, 2023{\natexlab{b}}.

\bibitem[Longpre et~al.(2023)Longpre, Hou, Vu, Webson, Chung, Tay, Zhou, Le,
  Zoph, Wei, et~al.]{longpre2023flan}
Longpre, S., Hou, L., Vu, T., Webson, A., Chung, H.~W., Tay, Y., Zhou, D., Le,
  Q.~V., Zoph, B., Wei, J., et~al.
\newblock The flan collection: Designing data and methods for effective
  instruction tuning.
\newblock \emph{arXiv preprint arXiv:2301.13688}, 2023.

\bibitem[Mihaylov et~al.(2018)Mihaylov, Clark, Khot, and
  Sabharwal]{mihaylov2018can}
Mihaylov, T., Clark, P., Khot, T., and Sabharwal, A.
\newblock Can a suit of armor conduct electricity? a new dataset for open book
  question answering.
\newblock In \emph{Proceedings of the 2018 Conference on Empirical Methods in
  Natural Language Processing}, pp.\  2381--2391, 2018.

\bibitem[Qin et~al.(2023)Qin, Zhang, Gong, Ding, Xu, and
  Liu]{qin2023distribution}
Qin, H., Zhang, X., Gong, R., Ding, Y., Xu, Y., and Liu, X.
\newblock Distribution-sensitive information retention for accurate binary
  neural network.
\newblock \emph{International Journal of Computer Vision}, 131\penalty0
  (1):\penalty0 26--47, 2023.

\bibitem[Sakaguchi et~al.(2021)Sakaguchi, Bras, Bhagavatula, and
  Choi]{sakaguchi2021winogrande}
Sakaguchi, K., Bras, R.~L., Bhagavatula, C., and Choi, Y.
\newblock Winogrande: An adversarial winograd schema challenge at scale.
\newblock \emph{Communications of the ACM}, 64\penalty0 (9):\penalty0 99--106,
  2021.

\bibitem[Shao et~al.(2023)Shao, Chen, Zhang, Xu, Zhao, Li, Zhang, Gao, Qiao,
  and Luo]{shao2023omniquant}
Shao, W., Chen, M., Zhang, Z., Xu, P., Zhao, L., Li, Z., Zhang, K., Gao, P.,
  Qiao, Y., and Luo, P.
\newblock Omniquant: Omnidirectionally calibrated quantization for large
  language models.
\newblock \emph{arXiv preprint arXiv:2308.13137}, 2023.

\bibitem[Sun et~al.(2023)Sun, Liu, Bair, and Kolter]{sun2023simple}
Sun, M., Liu, Z., Bair, A., and Kolter, J.~Z.
\newblock A simple and effective pruning approach for large language models.
\newblock \emph{arXiv preprint arXiv:2306.11695}, 2023.

\bibitem[Taori et~al.(2023)Taori, Gulrajani, Zhang, Dubois, Li, Guestrin,
  Liang, and Hashimoto]{taori2023stanford}
Taori, R., Gulrajani, I., Zhang, T., Dubois, Y., Li, X., Guestrin, C., Liang,
  P., and Hashimoto, T.~B.
\newblock Stanford alpaca: an instruction-following llama model (2023).
\newblock \emph{URL https://github. com/tatsu-lab/stanford\_alpaca}, 2023.

\bibitem[Touvron et~al.(2023{\natexlab{a}})Touvron, Lavril, Izacard, Martinet,
  Lachaux, Lacroix, Rozi{\`e}re, Goyal, Hambro, Azhar,
  et~al.]{touvron2023LLaMA}
Touvron, H., Lavril, T., Izacard, G., Martinet, X., Lachaux, M.-A., Lacroix,
  T., Rozi{\`e}re, B., Goyal, N., Hambro, E., Azhar, F., et~al.
\newblock Llama: Open and efficient foundation language models.
\newblock \emph{arXiv preprint arXiv:2302.13971}, 2023{\natexlab{a}}.

\bibitem[Touvron et~al.(2023{\natexlab{b}})Touvron, Martin, Stone, Albert,
  Almahairi, Babaei, Bashlykov, Batra, Bhargava, Bhosale,
  et~al.]{touvron2023LLaMA2}
Touvron, H., Martin, L., Stone, K., Albert, P., Almahairi, A., Babaei, Y.,
  Bashlykov, N., Batra, S., Bhargava, P., Bhosale, S., et~al.
\newblock Llama 2: Open foundation and fine-tuned chat models.
\newblock \emph{arXiv preprint arXiv:2307.09288}, 2023{\natexlab{b}}.

\bibitem[Wang et~al.(2022)Wang, Kordi, Mishra, Liu, Smith, Khashabi, and
  Hajishirzi]{wang2022self}
Wang, Y., Kordi, Y., Mishra, S., Liu, A., Smith, N.~A., Khashabi, D., and
  Hajishirzi, H.
\newblock Self-instruct: Aligning language model with self generated
  instructions.
\newblock \emph{arXiv preprint arXiv:2212.10560}, 2022.

\bibitem[Xiao et~al.(2023)Xiao, Lin, Seznec, Wu, Demouth, and
  Han]{xiao2023smoothquant}
Xiao, G., Lin, J., Seznec, M., Wu, H., Demouth, J., and Han, S.
\newblock Smoothquant: Accurate and efficient post-training quantization for
  large language models.
\newblock In \emph{International Conference on Machine Learning}, pp.\
  38087--38099. PMLR, 2023.

\bibitem[Xu et~al.(2023{\natexlab{a}})Xu, Xu, and Mandic]{xu2023tensorgpt}
Xu, M., Xu, Y.~L., and Mandic, D.~P.
\newblock Tensorgpt: Efficient compression of the embedding layer in llms based
  on the tensor-train decomposition.
\newblock \emph{arXiv preprint arXiv:2307.00526}, 2023{\natexlab{a}}.

\bibitem[Xu et~al.(2023{\natexlab{b}})Xu, Xie, Gu, Chen, Chang, Zhang, Chen,
  Zhang, and Tian]{xu2023qalora}
Xu, Y., Xie, L., Gu, X., Chen, X., Chang, H., Zhang, H., Chen, Z., Zhang, X.,
  and Tian, Q.
\newblock Qa-lora: Quantization-aware low-rank adaptation of large language
  models.
\newblock \emph{arXiv preprint arXiv:2309.14717}, 2023{\natexlab{b}}.

\bibitem[Zellers et~al.(2019)Zellers, Holtzman, Bisk, Farhadi, and
  Choi]{zellers2019hellaswag}
Zellers, R., Holtzman, A., Bisk, Y., Farhadi, A., and Choi, Y.
\newblock Hellaswag: Can a machine really finish your sentence?
\newblock In \emph{Proceedings of the 57th Annual Meeting of the Association
  for Computational Linguistics}, pp.\  4791--4800, 2019.

\bibitem[Zhang et~al.(2023)Zhang, Shen, Yang, Ou, Yu, Zhuang,
  et~al.]{zhang2023pruning}
Zhang, M., Shen, C., Yang, Z., Ou, L., Yu, X., Zhuang, B., et~al.
\newblock Pruning meets low-rank parameter-efficient fine-tuning.
\newblock \emph{arXiv preprint arXiv:2305.18403}, 2023.

\bibitem[Zhao et~al.(2023)Zhao, Zhou, Li, Tang, Wang, Hou, Min, Zhang, Zhang,
  Dong, et~al.]{zhao2023survey}
Zhao, W.~X., Zhou, K., Li, J., Tang, T., Wang, X., Hou, Y., Min, Y., Zhang, B.,
  Zhang, J., Dong, Z., et~al.
\newblock A survey of large language models.
\newblock \emph{arXiv preprint arXiv:2303.18223}, 2023.

\bibitem[Zhu et~al.(2023)Zhu, Li, Liu, Ma, and Wang]{zhu2023survey}
Zhu, X., Li, J., Liu, Y., Ma, C., and Wang, W.
\newblock A survey on model compression for large language models.
\newblock \emph{arXiv preprint arXiv:2308.07633}, 2023.

\end{thebibliography}
\bibliographystyle{icml2024}
}

\newpage
\appendix
\onecolumn

\section{Details about IR-QLoRA}

\subsection{Details about Related Works}
\label{sec:Details_about_Related_Works}

\textbf{Large Language Models}.
With the significant development of deep learning~\cite{tao2021towards,DBLP:conf/cvpr/WangLYLTL21,Zhao_2023_ICCV_DDFM,NAPGuard,tao2022exploring,Zhao_2024_CVPR,tao2022exploring,tripletsuppression,Zhao_2022_CVPR,wei2020occluded,Zhao_2023_CVPR,DBLP:journals/tip/WangLBL22,Zhao_2023_ICCV_Spherical}, LLMs have demonstrated remarkable proficiency across diverse natural language understanding tasks and are established as a prominent paradigm in this field~\cite{chang2023survey,devlin2018bert,zhao2023survey,huang2022towards}. Recent noteworthy instances of LLMs encompass OpenAI's GPT family~\cite{brown2020language} and Meta's LLaMA and LLaMA2 families~\cite{touvron2023LLaMA,touvron2023LLaMA2}. Nonetheless, the exceptional performance of these LLMs is contingent upon extensive parameters and computational resources. Notably, the PaLM-540B model boasts an impressive 540 billion parameters~\cite{chowdhery2023palm}, underscoring the substantial computational demands. This reality poses substantial challenges to deploying LLMs in settings with limited resources. Consequently, the research of the compression technologies for LLMs has gained prominence as a critical area of research.

\textbf{Compression of LLMs}.
Existing compression technologies of LLMs include pruning, distillation, low-rank decomposition, and low-bit quantization~\cite{ganesh2021compressing,zhu2023survey,chitty2023survey}. Pruning removes redundant parameters in LLMs structurally or unstructuredly~\cite{frantar2023sparsegpt,zhang2023pruning,sun2023simple}. Distillation enables the compressed student LLMs to learn from intermediate features or predictions of a larger teacher model~\cite{gu2023knowledge,huang2022context,agarwal2023gkd,li2022explanations,jiang2023lion}. Low-rank decomposition saves computation by decomposing the weight of LLMs into smaller matrices with significantly reduced dimensions~\cite{xu2023tensorgpt}. Different from the above technologies that mainly reduce the number of parameters (weights and/or activations), quantization aims to compress the LLMs from 16-bit floating-point to lower bit-widths to mitigate the storage and computation.

\textbf{Quantization of LLMs}.
Since compression is from a generic bit-width perspective, quantization has become a popular method to obtain efficient LLMs~\cite{frantar2022gptq,xiao2023smoothquant}. The most common practice is directly performing post-training quantization on pre-trained LLMs and optimizing quantizers through calibration~\cite{lin2023awq,lee2023owq,shao2023omniquant,dettmers2022llm}, which usually results in non-negligible degradation. Some quantization-aware training methods finetune the parameters of quantized LLMs to improve accuracy~\cite{liu2023llm,kim2023memory}, while the computational burden brought is significantly expensive. The LoRA-finetuning quantization of LLMs emerges to achieve a balanced trade-off between computational cost and accuracy~\cite{dettmers2023qlora,xu2023qalora,li2023loftq}, where LLMs are first quantized and then finetuned with a parameter-efficient LoRA. However, existing quantized LLMs with LoRA are still far from ideal in accuracy.

\subsection{Details about Pipeline}
\label{Details about Pipeline}
In this section, we demonstrate the application of \textit{Information Calibration Quantization} (ICQ) and \textit{Information Elastic Connection} (IEC) in our IR-QLoRA in detail when we quantizing the LLMs.

\begin{algorithm}[!ht]
\caption{The weight search process within each block in IR-QLoRA}\label{alg:ICQ}
\begin{algorithmic}
\STATE {\bfseries Input:} Block Weight $\boldsymbol{w}$, hypermeters $\lambda,\sigma,n$
\STATE {\bfseries Output:} Calibration constant $\tau_1^{\text{FP}8}, \tau_2^{\text{FP}16}$
\STATE Initialize $\tau_0=\operatorname{quantile}_{\frac{1}{2}}\left(\boldsymbol{w}\right), \mathcal{H}^* = 0$
\FOR{$\tau$ in $\operatorname{Linspace}(\tau_0 - \lambda\sigma, \tau_0 + \lambda\sigma, \frac{\lambda\sigma}{n})$} 
    \STATE $\boldsymbol{w} = \boldsymbol{w} - \tau$
    \STATE $\hat{\boldsymbol{w}} = \text{NF}{k}(\boldsymbol{w} / \operatorname{absmax}(\boldsymbol{w}) ) $
    \STATE Calculate the probability $P(q_i)$ of the $\hat{\boldsymbol{w}}$ taking the value $q_i$
    \STATE $\mathcal{H} = \sum_{i=0}^{2^k-1} P(q_i) \log_2{P(q_i)}  $
    \IF{$\mathcal{H} > \mathcal{H}^*$}
        \STATE Update $\tau^*$ and $\mathcal{H}^*$
    \ENDIF
\ENDFOR
\STATE $\tau_1^{\text{FP8}} = {\operatorname{FP}\!8}({\tau^*}/\tau_2^{\text{FP16}}) = {\operatorname{FP}\!8}({\tau^*}/\operatorname{absmax}(\tau^*))$ \qquad 
\end{algorithmic}
\end{algorithm}

\textbf{Pipeline of ICQ}.
The targets of ICQ are all weights that need to be quantized.
In the initial phase, these weights are partitioned into distinct blocks according to blocksize $B$.
Thereafter, an optimization process is engaged, aiming to identify a calibration constant that is optimal for maximizing the entropy of the quantized weights within each block.
The implementation details of the above functions are listed in Algorithm~\ref{alg:ICQ}. 

Note that the quantization of $\tau$ is a discretionary choice, contingent upon the preference for either elevated accuracy or diminished storage requirements. By default, we elect to implement the quantization of 
$\tau$ as a standard protocol.

Besides, in integer quantization like GPTQ~\cite{frantar2022gptq}, the ICQ strategy remains applicable for refining the entropy profile of quantized weights. This is actualized by the integration of the weights' entropy into the loss function during the calibration phase, thereby facilitating precision calibration.

\begin{algorithm}[!h]
\caption{The inference process of the linear layer in IR-QLoRA} \label{alg:IEC}
\begin{algorithmic}
\STATE {\bfseries Input:} Input tensor $\boldsymbol{x}\in\mathbb{R}^{b\times h}$, original weights $\boldsymbol{w}\in\mathbb{R}^{o\times h}$, low-rank paramaters $\boldsymbol{\ell}_1\in\mathbb{R}^{h\times r}, \boldsymbol{\ell}_2\in\mathbb{R}^{r\times o}$,  parameters $\alpha, \beta_1, \beta_2$
\STATE {\bfseries Output:} Output tensor $\boldsymbol{y}\in\mathbb{R}^{b\times o}$
\STATE $\hat{\boldsymbol{w}}_{\text{ICQ}}^{\text{NF}k}, s_1^{\text{FP8}}, s_2^{\text{FP16}}, \tau_1^{\text{FP8}}, \tau_2^{\text{FP16}} \gets $ search process for $\boldsymbol{w}$
\STATE $\hat{\boldsymbol{w}}_{\text{ICQ}}^\text{FP16}=\hat{\boldsymbol{w}}_{\text{ICQ}}^{\text{NF}k}\operatorname{dequant}(s_1^{\text{FP8}}, s_2^{\text{FP16}})+\operatorname{dequant}(\tau_1^{\text{FP8}}, \tau_2^{\text{FP16}}) $
\STATE $\boldsymbol{y}'_{\text{ICQ}} = \boldsymbol{x}\hat{\boldsymbol{w}}_{\text{ICQ}}^{\text{FP}16} $
\STATE $ \boldsymbol{x}' = U_1(\boldsymbol{x})=\boldsymbol{x}\boldsymbol{\ell}_1+\beta_1{\frac{r}{h}}\sum\limits_{i=1}^{{r}}\boldsymbol{x}^{\left[(i-1)\frac{h}{r} : i\frac{h}{r}-1\right]} $
\STATE $ U_{\text{IEC}}(\boldsymbol{x}) =  U_2(\boldsymbol{x}')=\boldsymbol{x}'\boldsymbol{\ell}_2+\beta_2\prod\limits^{{r}}{\boldsymbol{x}'}^{\left[0 : r-1\right]} $
\STATE $\boldsymbol{y} = \boldsymbol{y}'_{\text{ICQ}} + \alpha U_{\text{IEC}}(\boldsymbol{x}) $
\end{algorithmic}
\end{algorithm}

\textbf{Pipeline of IEC}.
The targets of IEC are specifically focused on the low-rank adaptations added in linear layers. We modified the forward function of LoRA, and the current implementation of the forward function is shown in Algorithm~\ref{alg:IEC}.

The parameters $\beta_1$ and $\beta_2$ introduced by the IEC can be integrated into the learned LoRA adapters to avoid additional inference costs. Specifically, by defining the matrix $\widetilde{\boldsymbol{\ell}_1}\in \mathbb{R}^{h\times r}$ and $\widetilde{\boldsymbol{\ell}_2}\in \mathbb{R}^{r\times o}$ as Eq.~\eqref{eq:IEC_merge_1}:

\begin{equation}
\label{eq:IEC_merge_1}
\begin{aligned}
\widetilde{\boldsymbol{\ell}_1}_{\{i, j\}} &=
\begin{cases}
    {\boldsymbol{\ell}_1}_{\{i, j\}} + \beta_1\frac{\operatorname{gcd}(h, r)}{h} &  \text{if } 
    \lfloor i / \frac{h}{\operatorname{gcd}(h, r)} \rfloor = \lfloor j /  \frac{r}{\operatorname{gcd}(h, r)} \rfloor, \\
    {\boldsymbol{\ell}_1}_{\{i, j\}} & \text{otherwise}, 
\end{cases} \\
\widetilde{\boldsymbol{\ell}_2}_{\{i, j\}} &=
\begin{cases}
    {\boldsymbol{\ell}_2}_{\{i, j\}} + \beta_2\frac{\operatorname{gcd}(o, r)}{r} &  \text{if } \lfloor i /  \frac{r}{\operatorname{gcd}(o, r)} \rfloor = \lfloor j /  \frac{o}{\operatorname{gcd}(o, r)} \rfloor, \\
    {\boldsymbol{\ell}_2}_{\{i, j\}} & \text{otherwise},
\end{cases} 
\end{aligned}
\end{equation}
where $\{i,j\}$ is the index of the matrix, and $\lfloor\cdot\rfloor$ denotes the floor operation.

The function $U_1(\boldsymbol{x})$ and $U_2(\boldsymbol{x})$ can be further redefined, as Eq.~\eqref{eq:IEC_merge_2}:

\begin{equation}
\label{eq:IEC_merge_2}
\begin{split}
U_1(\boldsymbol{x}) &= \boldsymbol{x}\widetilde{\boldsymbol{\ell}_1}
= \boldsymbol{x}\boldsymbol{\ell}_1 + \beta_1\prod\limits^{\frac{r}{\operatorname{gcd}{(h, r)}}}\left(\frac{\operatorname{gcd}{(h, r)}}{h}\sum\limits_{i=1}^{{\operatorname{gcd}{(h, r)}}}\boldsymbol{x}^{\left[(i-1)\frac{h}{\operatorname{gcd}{(h, r)}}: i\frac{h}{\operatorname{gcd}{(h, r)}}-1\right]}\right), \\
U_2(\boldsymbol{x}) &= \boldsymbol{x}\widetilde{\boldsymbol{\ell}_2}
= \boldsymbol{x}\boldsymbol{\ell}_2 + \beta_2\prod\limits^{\frac{o}{\operatorname{gcd}{(o, r)}}}\left(\frac{\operatorname{gcd}{(o, r)}}{r}\sum\limits_{i=1}^{{\operatorname{gcd}{(o, r)}}}{\boldsymbol{x}}^{\left[(i-1)\frac{r}{\operatorname{gcd}(o, r)}: i\frac{r}{\operatorname{gcd}(o, r)}-1\right]}\right). \\
\end{split}
\end{equation}

To summarize, ICQ and IEC within IR-QLoRA exhibit remarkable generality and both can be seamlessly integrated into any methods for LoRA-finetuning quantization of LLMs (\eg QLoRA~\cite{dettmers2023qlora}, QA-LoRA~\cite{xu2023qalora}). Moreover, their implementation incurs minimal overhead.

\clearpage

\section{Experiment Settings}
\label{sec:Detailed_Experiment_Settings}

\subsection{Models and Datasets}
For models, we establish IR-QLoRA upon the LLaMA~\cite{touvron2023LLaMA} and LLaMA2~\cite{touvron2023LLaMA2} families.
Specifically, we finetune the 7B, 13B, 30B, and 65B models of LLaMA and the 7B and 13B models of LLaMA2.

For datasets, we choose Alpaca~\cite{taori2023stanford} and FLAN v2~\cite{longpre2023flan} as our finetuning datasets.
Alpaca contains 52K instruction-following data generated from text-davinci-003 (GPT 3.5)~\cite{wang2022self}. 
Flan v2 is a collection of 1,836 tasks combining the mixture with CoT, Muffin, T0-SF, and NIV2.

\subsection{NormalFloat Quantization}
\label{sec:NormalFloat_Quantization}

Quantile Quantization~\cite{dettmers20218} is an information-theoretically optimal data type that ensures each quantization bin has an equal number of values from the input tensor. Essentially, it distributes the data evenly across the available quantization levels, leading to efficient and balanced utilization of the quantized representation.

Building on Quantile Quantization, NormalFloat (NF) Quantization~\cite{dettmers2023qlora} introduces the quantization principle that the weights conform to a zero-centered normal distribution.
By converting all weights into a fixed distribution, their distribution is fully adapted to the range of the specified data type.
Then the quantile constant can be calculated.
Following the processing steps in QLoRA~\cite{dettmers2023qlora}, we can get different quantized points according to different quantized bit widths. Table~\ref{tab:nf2}-\ref{tab:nf4} shows the NF quantization for 2-4 bits, respectively. Note that we use symmetrical settings in NF2 to prevent excessive deviation of information.

\begin{table}[!htb]
\centering
\caption{The exact values of the NormalFloat 2-bit (NF2) data type}
\setlength{\tabcolsep}{11.mm}
{
\label{tab:nf2}
\begin{tabular}{cc  cc}
\toprule
\textbf{Index} & \textbf{Value} & \textbf{Index} & \textbf{Value} \\
\midrule
0 & -1.0 & 2 & 0.2525685131549835 \\
1 & -0.25256848335266113 & 3 & 1.0 \\
\bottomrule
\end{tabular}
}
\end{table}
\begin{table}[!htb]
\centering
\caption{The exact values of the NormalFloat 3-bit (NF3) data type}
\setlength{\tabcolsep}{11.mm}
{
\label{tab:nf3}
\begin{tabular}{cc cc}
\toprule
\textbf{Index} & \textbf{Value} & \textbf{Index} & \textbf{Value} \\
\midrule
0 & -1.0 & 4 & 0.16093020141124725 \\
1 & -0.4786292016506195 & 5 & 0.33791524171829224 \\
2 & -0.217141792178154 & 6 & 0.5626170039176941 \\
3 & 0.0 & 7 & 1.0 \\
\bottomrule

\end{tabular}
}
\end{table}

\begin{table}[!htb]
\centering
\caption{The exact values of the NormalFloat 4-bit (NF4) data type}
\setlength{\tabcolsep}{11.mm}
{
\label{tab:nf4}
\begin{tabular}{cc cc}
\toprule
\textbf{Index} & \textbf{Value} & \textbf{Index} & \textbf{Value} \\
\midrule
0 & -1.0 & 8 & 0.07958029955625534 \\
1 & -0.6961928009986877 & 9 & 0.16093020141124725 \\
2 & -0.5250730514526367 & 10 & 0.24611230194568634 \\
3 & -0.39491748809814453 & 11 & 0.33791524171829224 \\
4 & -0.28444138169288635 & 12 & 0.44070982933044434 \\
5 & -0.18477343022823334 & 13 & 0.5626170039176941 \\
6 & -0.09105003625154495 & 14 & 0.7229568362236023 \\
7 & 0.0 & 15 & 1.0 \\
\bottomrule

\end{tabular}
}
\end{table}

\newcolumntype{M}[1]{>{\raggedright\arraybackslash}m{#1}}

\begin{table}[t]
\caption{Examples for each the evaluation datasets}
\label{tab:evaluation_example}
\centering
\begin{tabular}{|M{0.12\textwidth}|M{0.35\textwidth}M{0.35\textwidth}|M{0.08\textwidth}|}
\hline
\textbf{Dataset} & \textbf{Question} & ~ & \textbf{Answer} \\
\hline

\multirow{3}{*}{MMLU} &
\multicolumn{2}{M{0.7\textwidth}|}{
Which of the following factors is associated with a decreased risk of Alzheimer's?
} & \multirow{3}{*}{B} \\
~ & (A) Being African or Hispanic American & (B) Eating fish & ~ \\
~ & (C) A lower level of education & (D) Being married & ~ \\
\hline

\multirow{3}{*}{HellaSwag} &
\multicolumn{2}{M{0.7\textwidth}|}{
A man is sitting on a roof. he
} & \multirow{3}{*}{D} \\
~ & (A) is using wrap to wrap a pair of skis. & (B) is ripping level tiles off. & ~ \\
~ & (C) is holding a rubik's cube. & (D) starts pulling up roofing on a roof. & ~ \\
\hline

\multirow{5}{*}{PIQA} &
\multicolumn{2}{M{0.7\textwidth}|}{
How do I ready a guinea pig cage for it's new occupants?
} & \multirow{5}{*}{A} \\
~ & \multicolumn{2}{M{0.7\textwidth}|}{
(A) Provide the guinea pig with a cage full of a few inches of bedding made of ripped paper strips, you will also need to supply it with a water bottle and a food dish.
} & ~ \\
~ &  \multicolumn{2}{M{0.7\textwidth}|}{
(B) Provide the guinea pig with a cage full of a few inches of bedding made of ripped jeans material, you will also need to supply it with a water bottle and a food dish.
} & ~ \\
\hline

\multirow{2}{*}{WinoGrande} &
\multicolumn{2}{M{0.7\textwidth}|}{
John moved the couch from the garage to the backyard to create space. The \_ is small.
} & \multirow{2}{*}{A} \\
~ & (A) garage & (B) backyard & ~ \\
\hline

\multirow{5}{*}{ARC-Easy} &
\multicolumn{2}{M{0.7\textwidth}|}{
Which factor will most likely cause a person to develop a fever?
} & \multirow{3}{*}{B} \\
~ & (A) a leg muscle relaxing after exercise & (B) a bacterial population in the bloodstream & ~ \\
~ & (C) several viral particles on the skin & (D) carbohydrates being digested in the stomach & ~ \\
\hline

\multirow{3}{*}{ARC-Challenge} &
\multicolumn{2}{M{0.7\textwidth}|}{
George wants to warm his hands quickly by rubbing them. Which skin surface will produce the most heat?
} & \multirow{3}{*}{A} \\
~ & (A) dry palms & (B) wet palms & ~ \\
~ & (C) palms covered with oil & (D) palms covered with lotion & ~ \\
\hline

\multirow{2}{*}{BoolQ} &
\multicolumn{2}{M{0.7\textwidth}|}{
Phantom pain sensations are described as perceptions that an individual experiences relating to a limb or an organ that is not physically part of the body. Limb loss is a result of either removal by amputation or congenital limb deficiency. However, phantom limb sensations can also occur following nerve avulsion or spinal cord injury. Is pain experienced in a missing body part or paralyzed area?
} & \multirow{2}{*}{B} \\
~ & (A) True & (B) False & ~ \\
\hline

\multirow{3}{*}{OpenBookQA} &
\multicolumn{2}{M{0.7\textwidth}|}{
A magnet will stick to
} & \multirow{3}{*}{A} \\
~ & (A) a belt buckle & (B) a wooden table, & ~ \\
~ & (C) a plastic cup & (D) a paper plate & ~ \\
\hline

\end{tabular}
\end{table}

\subsection{Evaluation Metrics}
Following QLoRA~\cite{dettmers2023qlora} and QA-LoRA~\cite{xu2023qalora}, we evaluate language understanding capabilities of the LLMs on Massively Multitask Language Understanding (MMLU) benchmark~\cite{hendrycks2020measuring} and commonsense reasoning ability on several Common Sense QA datasets. 

MMLU includes 57 language tasks, including humanities (Hums.), STEM, social science, etc. We utilize the MMLU evaluation script and prompts as outlined in ~\cite{hendryckstest2021}.
Common Sense QA datasets includes HellaSwag~\cite{zellers2019hellaswag}, PIQA~\cite{bisk2020piqa}, WinoGrande~\cite{sakaguchi2021winogrande}, ARC~\cite{clark2018think}, BoolQ~\cite{clark2019boolq}, and OpenBookQA~\cite{mihaylov2018can}. We adopt lm-evalharness~\cite{eval-harness} to produce the Common Sense QA results. 
Table~\ref{tab:evaluation_example} provides examples of each evaluation dataset. Given the multiple-choice (MC) nature of these tasks, accuracy is selected as the primary metric for evaluation. 

\subsection{Implementation Details}
All experiments are conducted on Tesla A100 GPUs. Following~\cite{dettmers2023qlora}, we apply the double quantization mechanism, and set the block size is 64 for quantization and 256 for double quantization. Regarding LoRA parameters, we set $r=64$, $\alpha=16$, and LoRA dropout of 0.1 for models up to 13B and 0.05 for 33B and 65B models. We employ the paged AdamW optimizer with a beta2 value of 0.999, and a learning rate of 2e-4 for models up to 13B and 1e-4 for 33B and 65B models., limiting the maximum gradient norm to 0.3 and adopting a constant learning rate strategy. Fine-tuning was executed for 10,000 and 20,000 steps on the Alpaca and FLAN v2 datasets, respectively, utilizing a batch size 16. 

\section{Experiment Results}
\label{sec:Detailed_Experiment_Results}

\begin{figure}[t]
    \centering
    \subfigure[Query Linear Projection]{
        \begin{minipage}[t]{0.48\textwidth}
            \centering
            \includegraphics[width=1\linewidth]{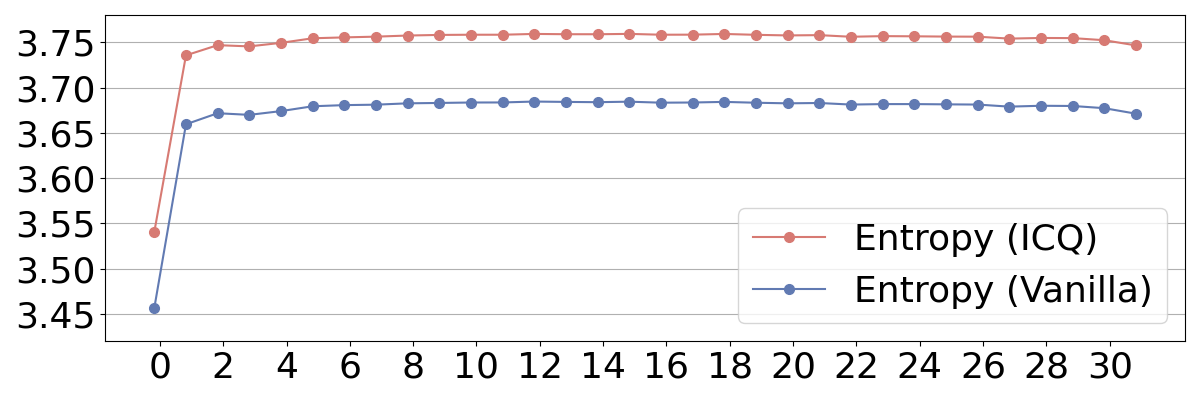}
        \end{minipage}
    }
    \subfigure[Value Linear Projection]{
        \begin{minipage}[t]{0.48\textwidth}
            \centering
            \includegraphics[width=1\linewidth]{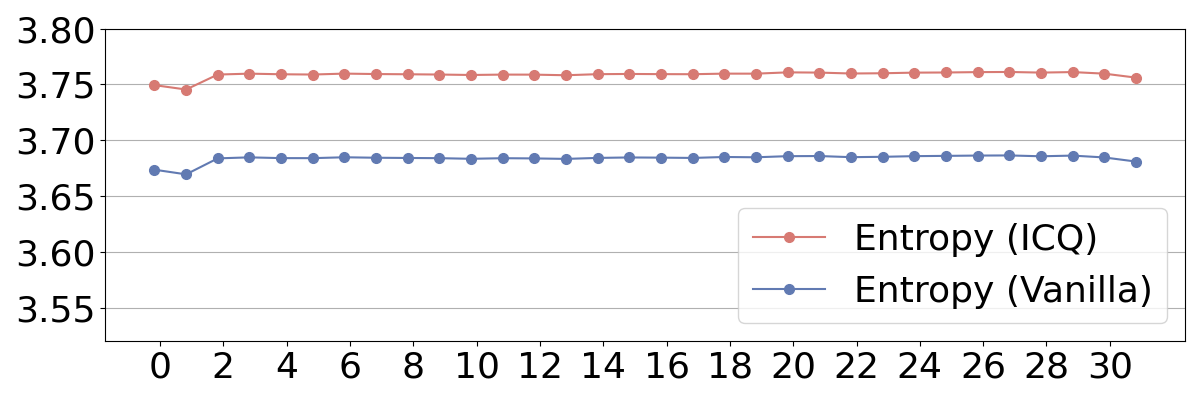}
        \end{minipage}
    }
    \subfigure[Output Linear Projection]{
        \begin{minipage}[t]{0.48\textwidth}
            \centering
            \includegraphics[width=1\linewidth]{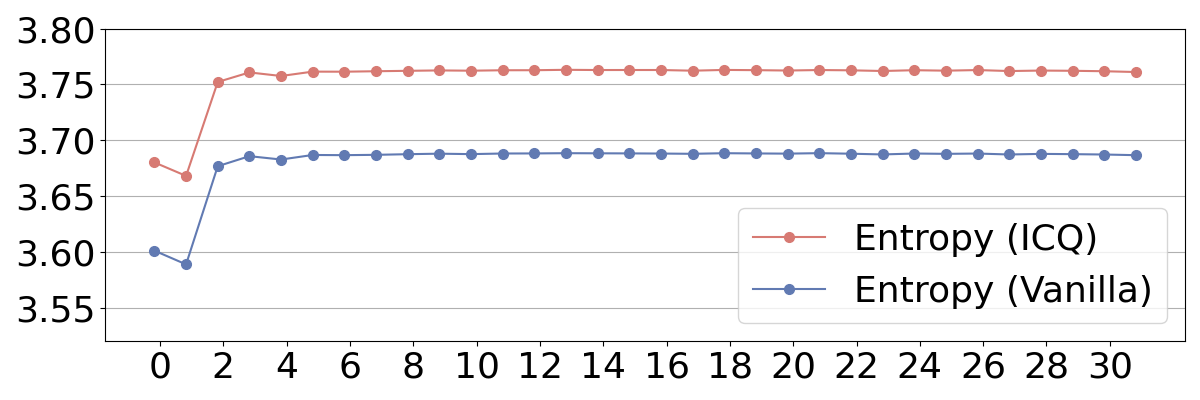}
        \end{minipage}
    }
    \subfigure[MLP Up Linear Projection]{
        \begin{minipage}[t]{0.48\textwidth}
            \centering
            \includegraphics[width=1\linewidth]{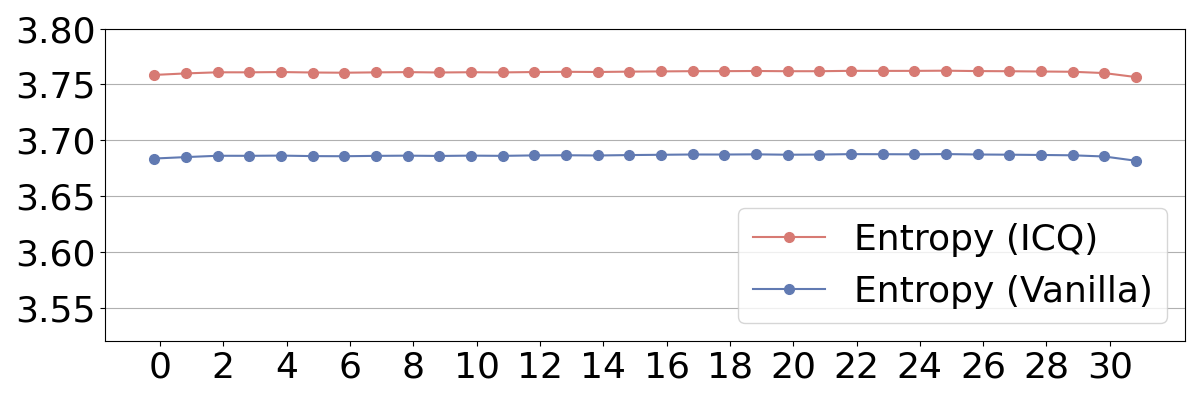}
        \end{minipage}
    }
    \subfigure[MLP Down Linear Projection]{
        \begin{minipage}[t]{0.48\textwidth}
            \centering
            \includegraphics[width=1\linewidth]{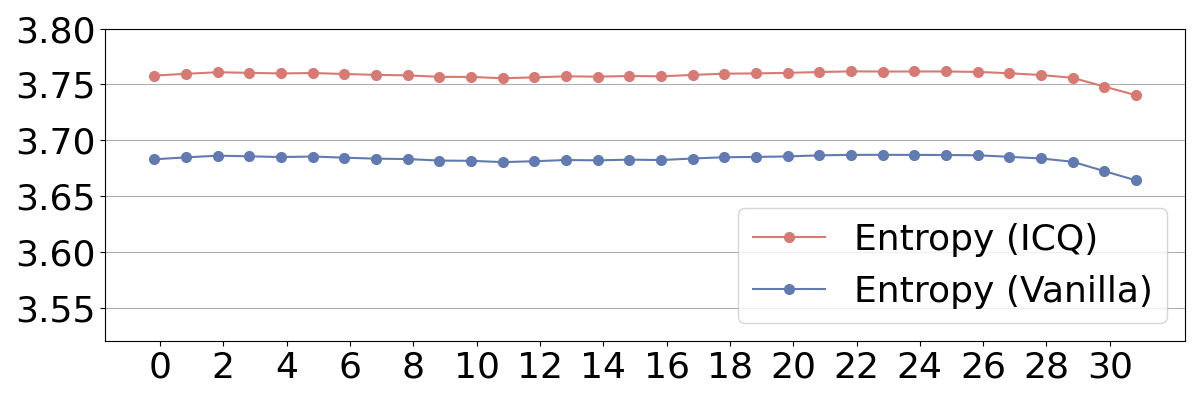}
        \end{minipage}
    }
    \subfigure[MLP Gate Linear Projection]{
        \begin{minipage}[t]{0.48\textwidth}
            \centering
            \includegraphics[width=1\linewidth]{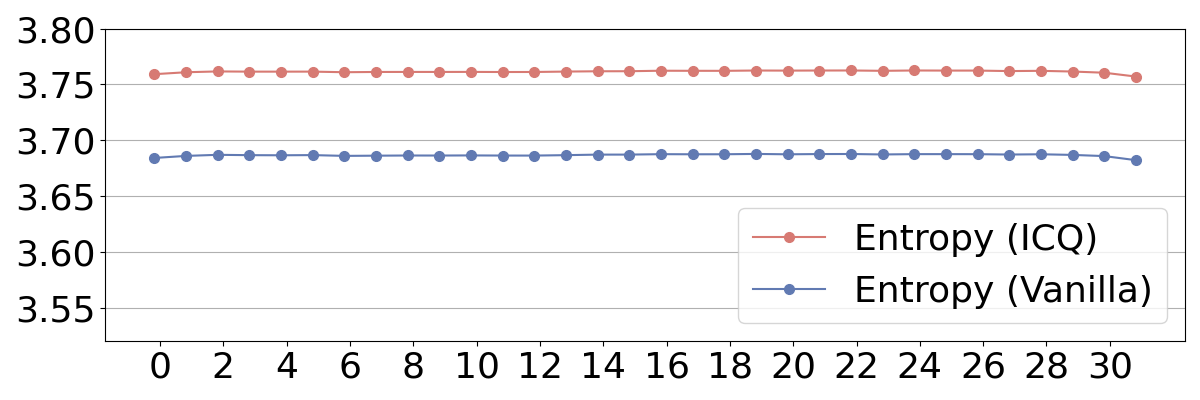}
        \end{minipage}
    }
    \caption{Entropy comparison of linear projections in 4-bit LLaMA-7B}
    \label{fig:Entropy}
\end{figure}

\subsection{Weight Visualization}

In addition to the linear projection for Key mentioned earlier, we also compute the entropy of quantized weights in various other types of layers, as illustrated in Figure~\ref{fig:Entropy}. This observation confirms that ICQ effectively boosts the information entropy of weights and augments the mutual information between the weights of quantized and original LLMs, consequently alleviating the constraints on representational capacity imposed by quantization.

\subsection{Efficiency Ablation}
We counted the number of parameters and training time for different sizes of LLaMA, as shown in Table~\ref{tab:ablation_params_all}. It is evident that IEC introduces a minimal number of additional parameters and does not increase the training time. In the case of ICQ, a small number of extra parameters are introduced, and the increase in training time is only incurred once, as the results can be efficiently cached after ICQ is applied. 
Thus, IR-QLoRA maintains nearly the same efficiency as Vanilla.

\begin{table}[htb]
    \footnotesize
    \centering
    \caption{Efficiency ablation on the different sizes of LLaMA}
    \label{tab:ablation_params_all}
    \setlength{\tabcolsep}{4.25mm}
    {
    \begin{tabular}{lc rrrrrrrr}
    \toprule
        \multirow{2}{*}{\textbf{Method}} & \multirow{2}{*}{\textbf{\#Bit}} & \multicolumn{4}{c}{\textbf{\textbf{\#Params}{\scriptsize (GB)}}} & \multicolumn{4}{c}{\textbf{Time{\scriptsize (h)}}} \\ \cmidrule(lr{.75em}){3-10} 
        ~ & ~ & \textbf{7B} & \textbf{13B} & \textbf{30B} & \textbf{65B}  & \textbf{7B} & \textbf{13B} & \textbf{30B} & \textbf{65B} \\
        \midrule
        LLaMA & 16 & 12.55 & 24.24 & 60.59 & 121.60 & - & - & - & -  \\
        \hdashline[0.5pt/2pt]
        Vanilla & 4 & 2.34 & 4.13 & 9.41 & 18.02 & 15.33 & 26.18 & 42.71 & 118.97 \\ 
        ICQ & 4 & 2.39 & 4.22 & 9.65 & 18.50 & 15.40 & 26.26 & 43.07 & 119.38 \\ 
        IEC & 4 & 2.34 & 4.13 & 9.41 & 18.02 & 15.33 & 26.18 & 42.71 & 118.97 \\ 
        \textbf{IR-QLoRA} & 4 & 2.39 & 4.22 & 9.65 & 18.50 & 15.40 & 26.26 & 43.07 & 119.38 \\ 
    \bottomrule
    \end{tabular}}
\end{table}

\definecolor{Red}{RGB}{220,20,60}
\definecolor{Blue}{RGB}{30,144,255}
\definecolor{Green}{RGB}{34,139,34}

\subsection{Case Study}
\label{sec:Case_Study}

This section offers a qualitative analysis that contrasts the performance of our IR-QLoRA model with that of QLoRA within the domain of LLM quantization. Utilizing prompts from the Stanford Alpaca dataset~\cite{taori2023stanford} for text generation, we indicate definitive answers in \textcolor{Blue}{blue} and unexpected responses in \textcolor{Red}{red} for clarity. Our qualitative assessment accentuates the enhanced capabilities of IR-QLoRA in intricate understanding and text generation quality, compared to the QLoRA baseline.

\textbf{Cognitive Reasoning}.
The cognitive reasoning aptitude of IR-QLoRA has shown notable advancement. When tasked with challenges that require logical deduction and computational acumen, such as those presented in arithmetic case studies and family relationship inference, as illustrated in Table~\ref{tab:case}, IR-QLoRA demonstrates a commendable level of proficiency. These results underscore its inherent reasoning capabilities and shed light on the potential for cognitive reasoning in quantized, smaller-scale models like QLoRA-7B.
Both IR-QLoRA-7B and IR-QLoRA-13B provided correct answers and demonstrated a capacity to explain their reasoning to a certain degree. In contrast, QLoRA-7B failed to answer correctly.

\textbf{Text Generation Quality}.
IR-QLoRA demonstrates substantial improvements in the coherence and fluency of generated text. When tasked with composing narratives for various scenarios, from mundane daily activities to complex extraterrestrial interactions, as cataloged in Table~\ref{tab:case}, IR-QLoRA invariably outshines QLoRA.
Both IR-QLoRA-7B and IR-QLoRA-13B demonstrate a remarkable ability to generate narratives with consistent themes and create more comprehensive stories. In contrast, QLoRA-7B often falters by misinterpreting the subject matter or producing narratives that lack completeness.

\begin{table}[t]
    \centering
    \renewcommand\arraystretch{1.}
    \caption{Case study for text generation outputs}
    \vspace{-0.1in}
    \label{tab:case}
\begin{tabular} {|M{0.2\textwidth}|M{0.75\textwidth}|} \hline

\multicolumn{2}{|M{0.95\textwidth}|}{
[1] \textbf{Input}: Alice has 15 apples. She buys 10 more from the market and then gives 7 apples to her friend. How many apples does Alice have now?}\\ \hline
\multicolumn{2}{|M{0.95\textwidth}|}{
\textcolor{Blue}{\textbf{Answer}: Alice have 18 apples now.}
}\\ \hline
 
\multicolumn{1}{|c|}{\textbf{Model}} & \multicolumn{1}{c|}{\textbf{Output}} \\ \hline
QLoRA-7B & 
\color{Red}{22}
\\ \hline
\textbf{IR-QLoRA-7B {\scriptsize (ours)}} & 
Alice now has 18 apples.
\\ \hline
QLoRA-13B & 
Alice had 15 apples. Alice buys 10 more apples. Alice has 15 + 10 = 25 apples. Alice gives 7 apples to her friend. Alice has 25 - 7 = 18 apples.
\\ \hline
\textbf{IR-QLoRA-13B {\scriptsize (ours)}} &  
Alice started with 15 apples and then bought 10 apples from the market, so her total apples inventory is now 25 apples (15 + 10 = 25). Then she gave 7 apples to her friend, so her total apples inventory is now 18 apples (25 - 7 = 18).
Alice now has 18 apples.
\\ \hline 
\multicolumn{2}{|M{0.95\textwidth}|}{
[2] \textbf{Input}: If Thomas is Joseph's son, what is Thomas to Joseph's father?} \\ \hline
\multicolumn{2}{|M{0.95\textwidth}|}{
\textcolor{Blue}{\textbf{Answer}: Thomas is Joseph's father's grandson }
}\\ \hline
\multicolumn{1}{|c|}{\textbf{Model}} & \multicolumn{1}{c|}{\textbf{Output}} \\ \hline
QLoRA-7B & 
\color{Red}{nephew}
 \\ \hline 
\textbf{IR-QLoRA-7B {\scriptsize (ours)}} & 
Thomas is grandson of Joseph's father.
\\ \hline
QLoRA-13B & 
grandson
\\ \hline
\textbf{IR-QLoRA-13B {\scriptsize (ours)}} &  
grandson
\\ \hline 
\multicolumn{2}{|M{0.95\textwidth}|}{
[3] \textbf{Input}: Imagine you are a historian living in ancient Rome, describe a day in your life.} \\ \hline
\multicolumn{1}{|c|}{\textbf{Model}} & \multicolumn{1}{c|}{\textbf{Output}} \\ \hline
QLoRA-7B & 
I wake up in my bed and eat breakfast. I put on my clothes and walk to school. I take my lunch and sit in my desk. \textcolor{Red}{We have a history lesson and a math lesson}. I walk home and play with my friends. \\ \hline
\textbf{IR-QLoRA-7B {\scriptsize (ours)}} & 
My day begins with waking up early to tend to my animals. {I then {head to the forum to listen to the debates between my fellow Romans}.} Afterwards, I might {head to the library to research some more}, before I finally return home to my family in the evening. \\ \hline
QLoRA-13B & 
\textcolor{Red}{You} wake up in your apartment in the center of the city. \textcolor{Red}{You} roll out of your bed and stretch your arms and legs. \textcolor{Red}{You} yawn and open your eyes. \textcolor{Red}{You} sit up and put your feet on the floor. \textcolor{Red}{You} get up and walk to the bathroom. \textcolor{Red}{You} wash your face, your hands and your body. \textcolor{Red}{You} then brush your teeth. \textcolor{Red}{You} return to your bedroom and get dressed in your clothes. \textcolor{Red}{You} then put on your shoes and walk out of your apartment. \\ \hline
\textbf{IR-QLoRA-13B {\scriptsize (ours)}} &  
As a historian living in ancient Rome, I would wake up in the morning when my slave would bring me coffee. Then I would go to the library and read ancient texts to see how Rome developed over time. Later, I would visit a slave market where I would buy my slaves. Then, I went to bed.
\\ \hline
\multicolumn{2}{|M{0.95\textwidth}|}{
[4] \textbf{Input}: Tell a story set on an alien planet, featuring an Earth astronaut and a representative of an alien civilization.} \\ \hline
\multicolumn{1}{|c|}{\textbf{Model}} & \multicolumn{1}{c|}{\textbf{Output}} \\ \hline
QLoRA-7B & 
This is an alien planet that has been invaded by an alien civilization.
 \\ \hline 
\textbf{IR-QLoRA-7B {\scriptsize (ours)}} & 
Earth astronaut John Smith landed on the alien planet of Tarsis. The planet Tarsis was inhabited by an alien civilization called the Ranax, who greeted Smith in a traditional Ranaxian ceremony. Smith and his Ranaxian counterpart, Kxat, bonded over their shared love of exploring space. Together, they set off on an adventure to uncover the mysteries of Tarsis, learning more about each other's cultures and planet along the way. \\ \hline
QLoRA-13B & 
As the spacecraft came to a stop on a distant planet, the astronauts got out and began to explore. \\ \hline
\textbf{IR-QLoRA-13B {\scriptsize (ours)}} &  
Earth astronaut Tom Jones landed on Mars and saw a Martian spaceship landing in front of him. Tom Jones approached the Martian spaceship and stepped inside. Tom Jones met a Martian representative named Martian Smith. Tom Jones asked Martian Smith "Are there any intelligent life forms on Mars?" Martian Smith replied "Yes, there are intelligent life forms on Mars." Tom Jones asked Martian Smith "How intelligent are these life forms?" Martian Smith replied "These life forms are very intelligent, they are more intelligent than any life form on Earth."
\\ \hline
\end{tabular}
\end{table}

\newpage
\appendix
\onecolumn

\end{document}